\documentclass[
twocolumn,
]{ceurart}

\usepackage{listings}
\usepackage{subcaption}

\begin{document}

\copyrightyear{2024}
\copyrightclause{Copyright for this paper by its authors.
  Use permitted under Creative Commons License Attribution 4.0
  International (CC BY 4.0).}

\conference{KI 2024: 47th German Conference on AI, 2nd Workshop on ‘Public Interest AI’,
  23 September, 2024, Würzburg, DE}

\title{Generating Synthetic Satellite Imagery for Rare Objects:\\An Empirical Comparison of Models and Metrics}


\author[1]{Tuong Vy Nguyen}[%
]
\fnmark[1]
\address[1]{Berlin University of Applied Sciences (BHT), Germany}

\author[1]{Johannes Hoster}[%
]
\fnmark[1]

\author[2,3]{Alexander Glaser}[%
]
\address[2]{Einstein Center Digital Future, Berlin, Germany}
\address[3]{Program on Science and Global Security, Princeton University, New Jersey, USA}

\author[1]{Kristian Hildebrand}[%
]

\author[1,2]{Felix Biessmann}[%
]
\cormark[1]

\cortext[1]{Corresponding author.}
\fntext[1]{These authors contributed equally.}

\begin{abstract}
  Generative deep learning architectures can produce realistic, high-resolution fake imagery -- with potentially drastic societal implications. Assessing the risks of this technology for the general public requires better understanding of the conditions under which novel generative methods can generate realistic data. A key question in this context is: How easy is it to generate realistic imagery, in particular for niche domains. The iterative process required to achieve specific image content is difficult to automate and control. Especially for rare classes, it remains difficult to assess \textit{fidelity}, meaning whether generative approaches produce realistic imagery and \textit{alignment}, meaning how (well) the generation can be guided by human input.
  In this work, we present a large-scale empirical evaluation of generative architectures which we fine-tuned to generate synthetic satellite imagery. We focus on nuclear power plants as an example of a rare object category - as there are only around 400 facilities worldwide, this restriction is exemplary for many other scenarios in which training and test data is limited by the restricted number of occurrences of real-world examples. We generate synthetic imagery by conditioning on two kinds of modalities, textual input and image input obtained from a game engine that allows for detailed specification of the building layout. The generated images are assessed by commonly used metrics for automatic evaluation and then compared with human judgement from our conducted user studies to assess their trustworthiness. Our results demonstrate that even for rare objects, generation of authentic synthetic satellite imagery with textual or detailed building layouts is feasible. However, in line with previous work, we find that automated metrics are often not aligned with human perception -- in fact, we find strong negative correlations between commonly used image quality metrics and human ratings. We believe that our findings enable researchers to better assess the strengths and weaknesses of different generative methods, especially for niche domains and rare object classes, and can help guide future improvements of generative methods.
\end{abstract}


\maketitle

\section{Introduction}

With the advent of novel generative methods for tabular data \cite{fonsecaTabularLatentSpace2023a,hassanDeepGenerativeModels2023}, text \cite{Zhao2021} and images \cite{rombachhighresolution2022}, synthetic data has entered the main stage of machine learning (ML) research. The applications are manifold, ranging from arts over software development to improving ML itself. 

For generative Artificial Intelligence (genAI), methods designed for text data, the risks and societal impact have been studied, for instance, in the context of large election campaigns \cite{centerforsecurityandemergingtechnologyTruthLiesAutomation2021}. For the image domain however, research on the technical underpinnings of the risks for the general public inherent to genAI technology have been underrepresented in the literature. 

\begin{figure}[t]
    \centering
    \includegraphics[width=0.99\columnwidth]{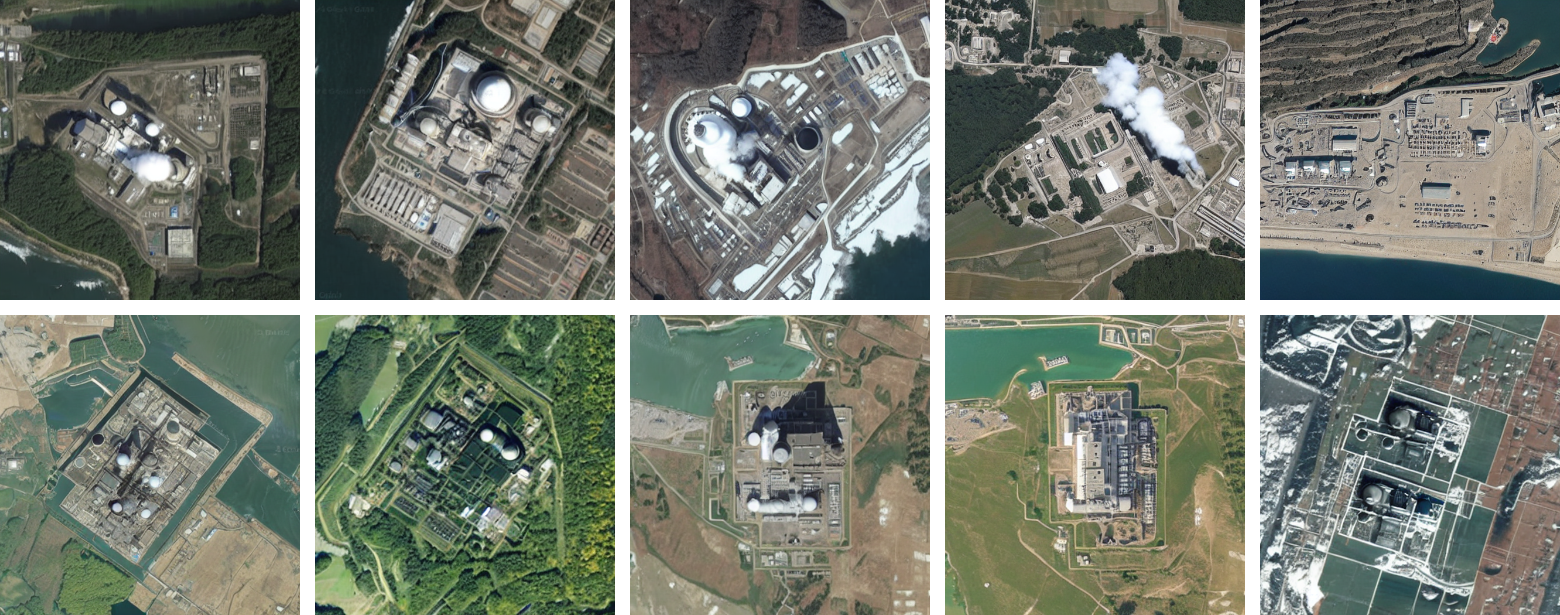}
    \caption[Synthetic samples.]{Samples of synthetic satellite imagery of facilities. Generated using fine-tuned generative models with only text input (top row) and the same fine-tuned models with additional image input (bottom row).}
    \label{fig:selected-syn-samples}
    \vspace{-1.0em}
\end{figure}

Within the ML community, the primary use case for synthetic data is arguably the generation of new training data for the development of larger and more powerful generative ML models. This application scenario has attracted attention, especially in the context of tech companies' demand for more data. These companies could soon run out of data for training language models \cite{cademetzHowTechGiants}. Synthetic data appears to be the solution to these problems of data-hungry large ML models, not only for these applications. Moreover, in other application domains, such as health care, synthetic data has attracted attention for different reasons: patient records are sensitive data which must not be shared publicly, hence synthetic patient record data could solve the problems around training ML models for healthcare without risking patients' privacy \cite{chenSyntheticDataMachine2021}. 

While the opportunities synthetic data offers are certainly convincing, there are two key challenges: For one, it remains difficult to control the output of these methods, second, it remains difficult to automatically assess the quality of generated data. In user-facing products, these two challenges are usually met with extensive human work -- there are highly-paid prompt engineering positions and crowd working platforms for evaluating the quality of generated data. These measures jeopardize the very idea of generative methods, automated data generation, and highlight the need for a better understanding of control and evaluation of generative methods.  

In this work, we investigate two main questions around state-of-the-art image generation methods: 1) \textit{How easily can they be controlled?} and 2) \textit{How trustworthy are established automatic evaluation metrics?} Our main contribution is a large-scale empirical evaluation with more than 30,000 human ratings of generated images. In contrast to previous work, we take the evaluated methods outside of their comfort zone: First we consider a niche domain, satellite imagery and second, we focus on a specific class of objects for which only a few hundred real-world instances exist, nuclear plants. 

\paragraph{Objective}

We aim to leverage and investigate ML techniques in combination with a pre-trained text-to-image model to generate synthetic satellite images for remote sensing purposes that could combat the issue of insufficient labeled imagery. With our approach, appropriate data could be easily generated in a controllable setting while maintaining low costs. To examine the generalizing capabilities of the pre-trained model and training on limited data, we generate imagery of a unique and rare class. Given the context, we opt for nuclear power plants as our target object and show how synthetic satellite imagery of a specific class can be created under set conditions. We use text prompts and additional image input as guidance for the generation process, and combine this workflow with fine-tuned models.

Moreover, we conduct an empirical study regarding the comparison of different approaches for conditional image synthesis, as well as the automated evaluation metrics used to assess the generated data. Current established evaluation metrics for the assessment of perceptual image quality are known to not necessarily align with human perception and are, thus, not always a reliable measure of image quality \cite{zhou2019hype}. In this work, we acquire quantitative evaluation results through larger-scale human-in-the-loop experiments, which we then compare to established metrics in the field.
To the best of our knowledge, such an in-depth empirical comparison between models, metrics and human ratings has not been done for rare object classes, such as nuclear facilities, in this manner before.

\section{Related Work}

In the following section we provide an overview of the current state of research in the relevant fields: Generative modeling, particularly text-to-image synthesis, deep learning (DL) applications in remote sensing, and evaluation methods in the context of genAI.

\subsection{Deep Generative Models for Image Synthesis}

With advances in ML and DL, the process of image generation can be automated and provides a way to generate data at relatively low cost. There are several popular model architectures that aim to generate data and for a long time, the state of the art in generative modeling have been Generative Adversarial Networks (GAN) \cite{goodfellow_gan_2014}, which have been explored thoroughly and applied in various domains and used as default for text-to-image translation over the years \cite{shamsolmoaliimage2020}. More recently, novel approaches to image generation based on Diffusion Models (DM) \cite{ho_denoising_2020} overcame some of the challenges associated with GAN training, with e.g. Dhariwal and Nichol \cite{dhariwaldiffusion2021} showing that DMs could outperform GANs in image synthesis. Currently, many generative model architectures, especially text-to-image ones, are based on DMs.

Text-to-image generation is a specific type of generative modeling which combines technologies of two different fields: Computer Vision and Natural Language Processing. Image generation conditioned on text has the advantage that it is very intuitive and easily comprehensible. There exist a variety of such recently developed vision-language models \cite{nicholglide2022, ramesh2022hierarchical, rameshzeroshot2021, rombachhighresolution2022, sahariaphotorealistic2022}, that have already shown how powerful such large text-to-image architectures can be. Especially their zero-shot ability - synthesizing images of concepts not seen during training - makes these models very compelling.

Furthermore, there are attempts to directly manipulate features in the latent space to create desired images. For example, by modelling the independent latent characteristics of an object through disentangled representations so that these features can be edited, e.g. changing pose and appearance respectively \cite{Esser_2019_ICCV}. Or by identifying latent directions through PCA, which enables to control GAN model-based image features like viewpoint, aging, lighting, and time of day \cite{NEURIPS2020_6fe43269}. Park et al. identify a local latent subspace within the latent space of a diffusion model, which enables image editing capabilities through movement along the basis vector at specific timesteps \cite{NEURIPS2023_4bfcebed}.

\subsection{Deep Learning in Remote Sensing}

There are a variety of use cases for the utilization of DL approaches in the context of satellite imagery and remote sensing, which, since 2014, several works have already dedicated themselves to \cite{ma_deep_2019}. Popular use-cases are scene classification, object detection and segmentation \cite{ma_deep_2019}.
With the emerging of generative models, GANs have found their way into the domain as well, mostly dealing with image-to-image translation tasks, e.g. translating city styles or creating cartographic representations from satellite images \cite{xu_satellite_2018, Zhao2021}. Tasks such as cloud removal and super-resolution are frequent use-cases as well.

However, there have been few works addressing the synthesis of novel imagery: \cite{yatesevaluation2022} have implemented and evaluated GANs for the synthesis of aerial imagery, but in an unconditional manner. Others have used special software tools instead of GANs with focus on certain objects and tasks like airplanes \cite{RarePlanes} and synthetic overhead imagery suitable for building segmentation \cite{Synthinel-1}. 
While most research aims to generate natural images, some works have already brought the text-to-image approach into the remote sensing domain, enabling the generation of different remote sensing images based on text descriptions \cite{bejigaretroremote2019,chenremoteaugm2021,zhaotexttoremote2022, xutxt2imgmhn2022}. However, these were somewhat restricted by the limited amount of suitable image-text datasets and produced results leaving room for improvement. This work aims to further investigate the conditional generation of satellite imagery of a specific and rare class based on text descriptions by using a pre-trained text-to-image model and limited training data for remote sensing, which, to the best of our knowledge, has only been done in \cite{nguyen2023generating} so far. Moreover, the additional use of image input next to text prompts for further conditioning during the image synthesis process for satellite imagery, as done in \cite{hoster2023using}, is underrepresented as well.

\subsection{Evaluation of Generative AI}

For genAI, there are several ways to evaluate models and their outputs \cite{bandi_generativeai_2023}. The most common methods are described in the following:  

\paragraph{Automatic metrics.} The qualitative evaluation of generative models can be very subjective, and requires adequate quantitative metrics for the systematic assessment of generative models and their synthesized data. However, the evaluation of generative models, or their synthetic data, remains a challenging task: There are no standardized benchmarks or protocols set in place \cite{man_review_2022, otanitoward2023}, and especially in the domain of satellite imagery, finding suitable datasets for training as well as evaluation is rather difficult, particularly for the specific use case at hand. A lot of evaluation metrics, such as the commonly used Fr\'{e}chet Inception Distance (FID) \cite{heuselgans2018}, require sufficient real data for comparative analysis, which - like in our case - can be difficult to acquire, due to the fact alone that there are only a few hundred existing nuclear facilities in the world. The Inception Score (IS) \cite{salimansimproved2016} only assesses the synthetic images, which, albeit lacking a comparison to real data, is more practicable in our case. Since most metrics rely on the feature space of a pre-trained classification model, this limits the metrics to what the model knows, and renders them biased towards the used Inception model and, thus, the ImageNet dataset which the model was trained on \cite{stein2023exposing}. Another drawback is the need for a large sample size (typically around 50k) to make the metrics robust and reliable, which is not always feasible. Furthermore, the FID is known to not necessarily align with human visual perception \cite{zhou2019hype}, especially in the remote sensing and earth observation domain \cite{yatesevaluation2022}. The automatic, reliable evaluation of generative models and its outputs remains an ongoing research field \cite{bandi_generativeai_2023}.

\paragraph{Downstream tasks.} Another way of evaluating synthetic data is its use in a downstream application task: The generated images can be used in e.g. a classification task to observe how well they are classified by a pre-trained classifier \cite{chambon_adapting_2022}. A second method is to train a classification model on the synthetic data, or parts of it, and then apply the trained model on real unseen data and evaluate based on the predictive performance \cite{bandi_generativeai_2023}. However, for this method there has to be suitable real data to test with, which, like in our case, is not necessarily given.

\paragraph{Human evaluation.} Another method to assess genAI is human evaluation, with many works resorting to user studies to judge their synthetic data \cite{ding2022cogview2, sahariaphotorealistic2022, nicholglide2022, ramesh2022hierarchical, salimansimproved2016}. So-called human-in-the-loop experiments can be a valuable alternative when the aforementioned methods are not applicable or unreliable, and their results are easily comprehensible. Human ratings can give a more accurate assessment for specific tasks when automated metrics fail to reliably capture the image quality. Although there have been works proposing more standardized guidelines for evaluation \cite{zhou2019hype, otanitoward2023, Funke_2021}, there are no established protocols set in place for human experiments in genAI, which makes a comparison between published works quite difficult. Moreover, conducting human experiments requires additional work and resources, which are not always at disposal. We aim to conduct human evaluation for a use case, where offline metrics are likely unfitting, and follow recommendations regarding user study settings, to make our findings transparent.

\section{Method}\label{sec:method}

\begin{figure}[bht]
    \centering
    \includegraphics[width=0.99\columnwidth]{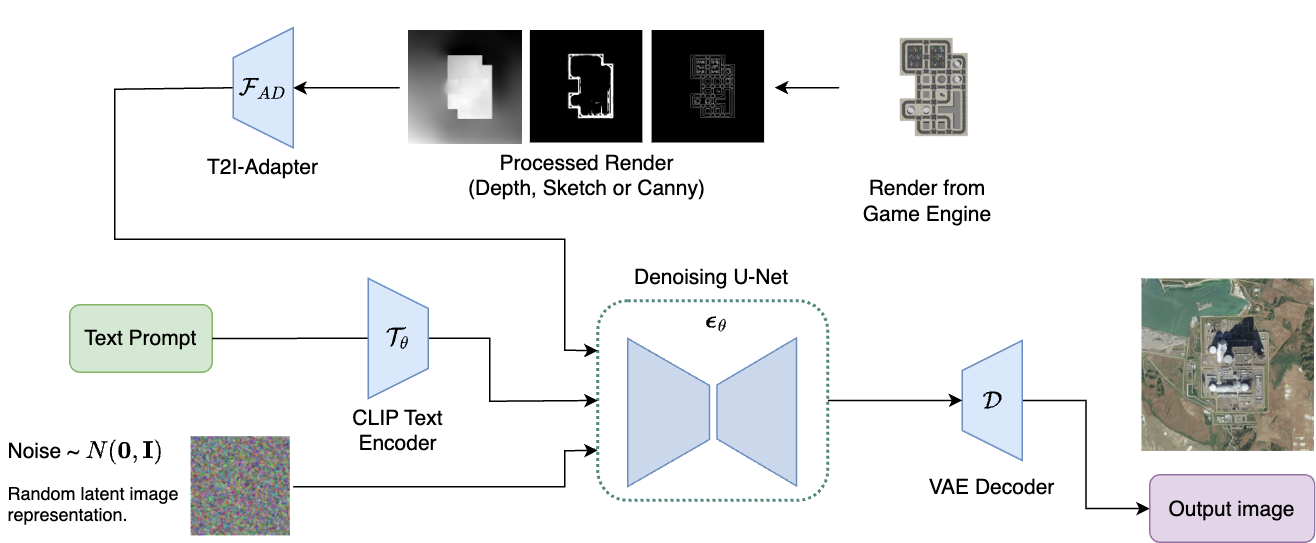}
    \caption[Workflow.]{Workflow of our image generation process. We use the renders from the game engine and process them into the respective input modalities (depth/sketch/canny). If used, they're put into the T2I-Adapter and used as structural guidance in the denoising component, together with the text embeddings obtained from the CLIP text encoder. A text prompt could be \textit{"an aerial view of a [*] nuclear power plant"}. All components are pre-trained, in case of fine-tuning, the U-Net (with DreamBooth) or text encoder (with Textual Inversion) are modified.}
    \label{fig:workflow}
    \vspace{-1.0em}
\end{figure}

By utilizing a pre-trained text-to-image model such as Stable Diffusion and fine-tuning techniques, we are able to leverage its prior knowledge while simultaneously adapting the model to our domain. We have fine-tuned this model on imagery of our target object, nuclear facilities, 
using DreamBooth \cite{ruizdreambooth2022} and Textual Inversion \cite{galimage2022} as fine-tuning methods. For more details, we refer to the respective sources.
To test the generalizability and the model's zero-shot capabilities, we use the unmodified pre-trained model as a baseline to examine how well the prior knowledge can be leveraged to generate satellite imagery of our target object. We then apply the mentioned fine-tuning approaches to further train the model on the datasets described below.
For further control during the generation process, we use additional conditioning input with the T2I-Adapter model \cite{mou2023t2iadapter,hoster2023using}.
Moreover, we combine the two approaches: Instead of using the original Stable Diffusion model as base for the T2I-Adapter, we exchange it with our fine-tuned versions (see \autoref{fig:workflow}).
The intuition is to leverage the newly learned concepts and use them in synergy with the additional image input. The model might then be more familiar with the given layout and able to generate data that better represents our desired image content. 

We have three approaches that rely only on text prompts as input and then all three methods used in combination with the T2I-Adapter, leaving us, in total, with six models to evaluate:
\begin{itemize}
    \item[] (1) the original unmodified Stable Diffusion model, \textit{SDiff T2I}, 
    \item[] (2) the DreamBooth fine-tuned model, \textit{DB T2I},
    \item[] (3) the fine-tuned one using Textual Inversion, \textit{TI T2I}, 
    \item[] (4) the base model (1) with the T2I-Adapter, \textit{SDiff T2I+Adapter}, 
    \item[] (5) the model from (2) with the T2I-Adapter, \textit{DB T2I+Adapter},
    \item[] (6) the model from (3) with the T2I-Adapter, \textit{TI T2I+Adapter}.
\end{itemize}

The implementation in this work relies heavily on Hugging Face's Diffusers Library \cite{huggingface_diffusers}. For all approaches, we use the publicly available Stable Diffusion v1.5 as base, a pre-trained vision-language model build on Latent Diffusion Models \cite{rombachhighresolution2022}. This version is compatible with the pre-trained T2I-Adapter components and lays a consistent foundation across methods for later comparison.

\subsection{Data}\label{sec:data}

To acquire data to train with, Google Earth Engine and web scraping tools are applied to obtain satellite and aerial imagery of nuclear facilities. After removing images where sites were
blurred or of low quality, the resulting dataset contains 202 satellite images of 185 unique nuclear power plants around the world. To exploit the model's prior, we apply conditionings - which are not present in the mentioned training data - to those newly learned concepts by adding keywords to the text prompts for variations regarding the location, seasonality and the time of day, for example, generating images of a nuclear facility in the desert or in the winter. Using different settings, synthetic images are generated for each approach.
For the additional image input, we use the T2I-Adapter \cite{mou2023t2iadapter} as described in \cite{hoster2023using}. We generate three different layouts of fictional power plants using the game engine Unity, varying the angle and rotation from which the site is looked at. This creates different viewpoints from the same facility. The renders are then turned into canny edge, depth maps and sketches for further structural guidance during the generation process (see \autoref{fig:workflow}). Images are then generated using layout conditioning in addition to the text prompts.

We generate a pool of images for each model, using different variations in the given text prompts. For the additional usage of the T2I-Adapter, we use different input modalities (canny, depth map, sketch) and vary the viewpoints for each of the three layouts. This way, we generate a variety of synthetic imagery, but based on the same three layouts originally rendered from the game engine. For the human experiments, we randomly select 500 images for each approach.
\autoref{fig:selected-syn-samples} shows samples of synthetic satellite imagery of nuclear facilities which have been generated using the methods mentioned previously in this section. For these, either a single text prompt (e.g. \textit{``an aerial view of a [*] nuclear power plant, forest, green''}) or a text prompt with an additional image have been used as input. For comparison, we also include the 202 real images in our human evaluation experiments. All images have been scaled to the same pixel size (512x512).

\subsection{Experiments}

To evaluate our generated data, we conduct a user study where we assess based on three aspects: (a) Fidelity (image quality), how authentic does an image look, (b) text alignment (semantic control), how well does an image match the given text prompt and (c) layout alignment (structural control), how well can the structure of the same subject be retained within several images. For human experiments in genAI, some works opt for a 2-alternative forced choice setup \cite{rameshzeroshot2021, sahariaphotorealistic2022, yatesevaluation2022}: Two images of two models are put next to each other and the user is tasked with selecting the superior one. However, this user interface design limits the comparison to only two options at a time. Similar to other work \cite{otanitoward2023, xu2023imagereward}, we apply a Likert scale where users are tasked to rate a given image or group of images from 1-5, depending on what aspect is being evaluated. This way, images and methods can be rated independently of one another and then later evaluated and ranked. 

\begin{figure}
\centering
    \begin{subfigure}{0.4\columnwidth}
        \centering
        \includegraphics[width=0.9\textwidth]{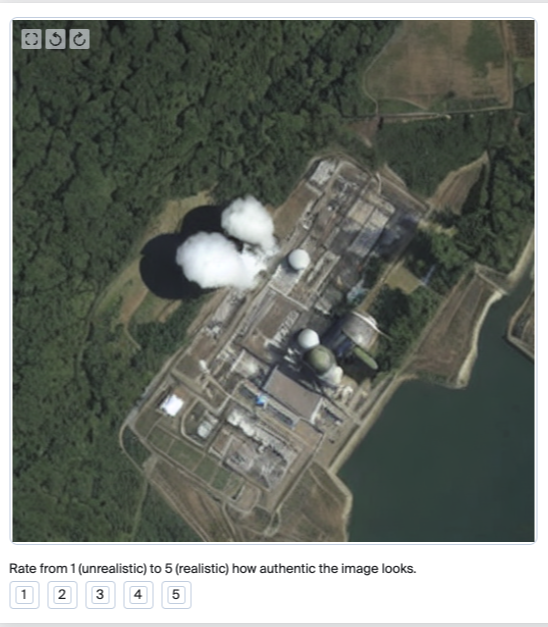}
        \caption{Image fidelity.}
        \label{fig:ui-fidelity}
    \end{subfigure}
    \begin{subfigure}{0.4\columnwidth}
        \centering
        \includegraphics[width=0.9\textwidth]{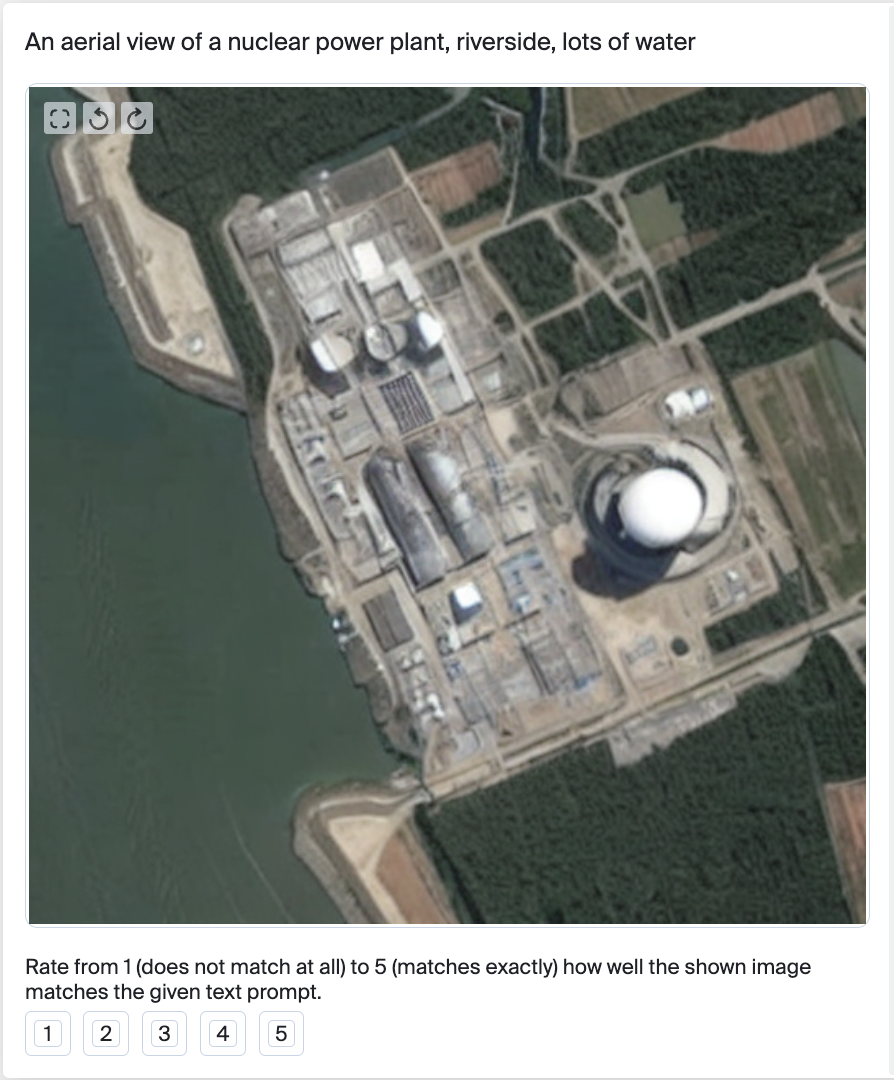}
        \caption{Text alignment.}
        \label{fig:ui-text}
    \end{subfigure}
    \begin{subfigure}{\columnwidth}
        \centering
        \includegraphics[width=0.75\columnwidth]{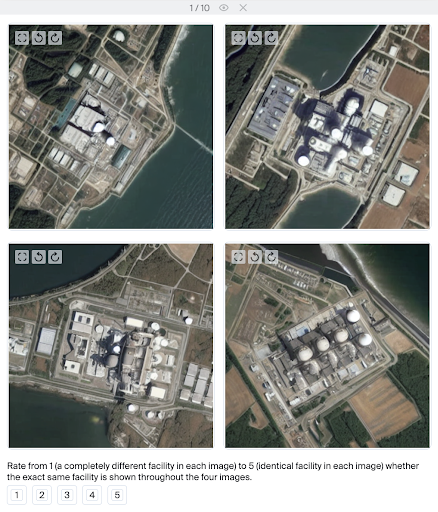}
        \caption{Layout alignment.}
        \label{fig:ui-layout}
    \end{subfigure}
    \caption{User interfaces of the conducted user studies. Shown are the study designs of the (a) image fidelity, (b) text alignment and (c) layout alignment assessments.}
    \label{fig:ui-examples}
    \vspace{-1.0em}
\end{figure}

\paragraph{User study.} We use the crowdsourcing platform Toloka \cite{toloka} and considered two main principles in the experimental design: The task should be simple and the results interpretable \cite{otanitoward2023}. The implemented user interface design for each study is shown in \autoref{fig:ui-examples}: For the (a) image fidelity analysis (see \autoref{fig:ui-fidelity}), users are instructed to rate a given image from 1 (unrealistic) to 5 (realistic) based on how authentic it looks to them. The interface for the (b) text alignment studies (see \autoref{fig:ui-text}) looks almost the same, with the exception that the text prompt that was used as conditioning input, is also shown. Participants have to rate from 1 (does not match at all) to 5 (matches exactly) how well the shown image matches the text. The third user study examines the (c) layout alignment. As depicted in \autoref{fig:ui-layout}, the user is shown four images that were generated from the same model, and asked to rate from 1 (completely different facility in each image) to 5 (identical facility) whether the shown images depict the same facility.
Participants were selected to be fluent in English and instructed about the motivation of the study; all participants confirmed acceptance of the data usage. We excluded responses from participants that submitted incomplete tasks, tasks in which users pressed only one key repeatedly and tasks in which control tasks (for which ground truth data was available) were answered incorrectly. 
For tasks (a) and (b) every participant rated 30 images in total. In task (c) each user rated 40 images.
Compensation was according to the minimum wage in the most frequent countries of origin on the platform.

\paragraph{Evaluation setup.}\label{sec:eval-setup}
Since we lack sufficient real data of our target object, as it is a rare object class, we only consider metrics which do not rely on the feature space of real data. Calculating scores like the FID \cite{heuselgans2018} with the 202 images (see Section \ref{sec:data}) might deliver unreliable results due to the low sample size and possibly contain bias, as these were already used for fine-tuning. Therefore, we only use the IS \cite{salimansimproved2016} as automatic metric in our evaluation, although it also is a flawed metric \cite{barratt_note_2018}. For implementation, we use torch-fidelity \cite{obukhov2020torchfidelity} to calculate the score. The IS is calculated as follows:
\begin{equation} \label{eq:inception-score}
  \text{IS} = \text{exp}(\ \mathbb{E}_{\boldsymbol{x}\sim p_\theta}[\ D_{KL}(p(y|\boldsymbol{x})\|p(y))]\ ). 
\end{equation} 
$\boldsymbol{x}$ is sampled from $p_\theta$, the encoded distribution of our synthetic images. The metric makes use of the KL divergence, calculated between the conditional label distribution $p(y|\boldsymbol{x})$ (favoring low entropy) and the marginal distribution $p(y)$ from all samples (favoring high entropy). For more details see \cite{salimansimproved2016}. To gain a score that might possibly be more accustomed to the remote sensing domain, we further apply an adapted version of the IS, as done in \cite{nguyen2023generating}: We exchange the pre-trained Inception model with a classifier fine-tuned on a land-use classification dataset \cite{yi_ucmerced_2010}, aerial-view imagery to give information on land cover. The modified IS is denoted as ${\text{IS}\textsubscript{adapt.}}$.

\begin{figure*}[ht!]
    \begin{subfigure}{0.33\textwidth}
        \centering
        \includegraphics[width=\textwidth]{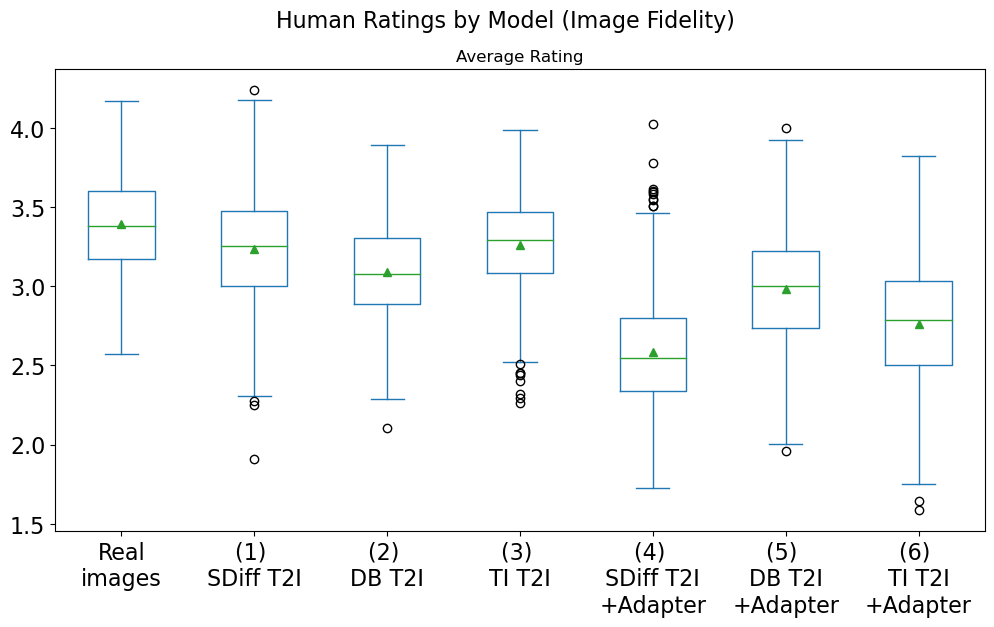}
        \caption{Image fidelity user study results.}
        \label{fig:boxplot-fidelity}
    \end{subfigure}
    \begin{subfigure}{0.33\textwidth}
        \centering
        \includegraphics[width=\textwidth]{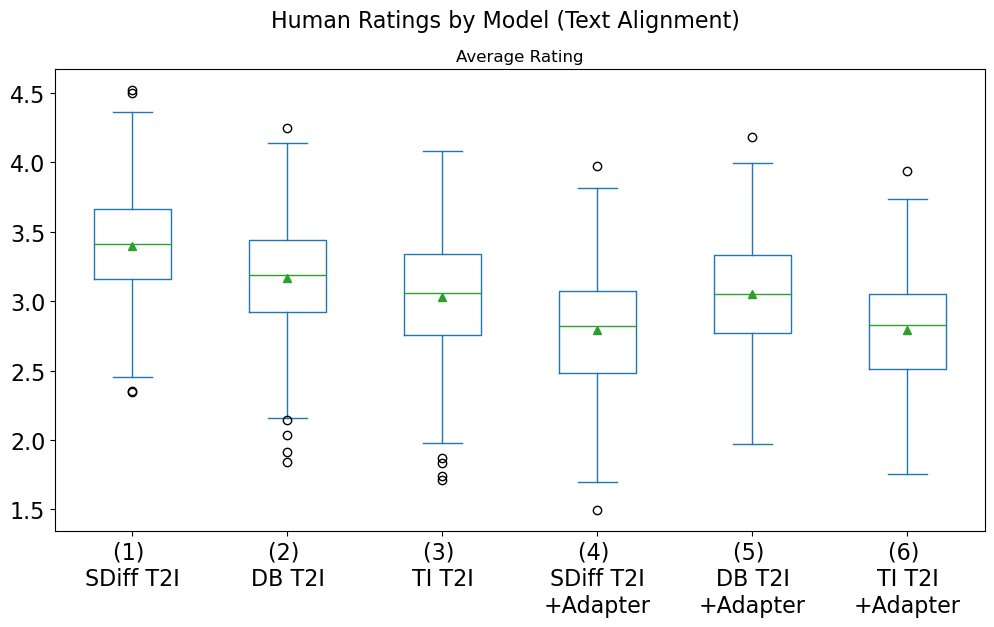}
        \caption{Text alignment user study results.}
        \label{fig:boxplot-text}
    \end{subfigure}
    \begin{subfigure}{0.33\textwidth}
        \centering
        \includegraphics[width=\textwidth]{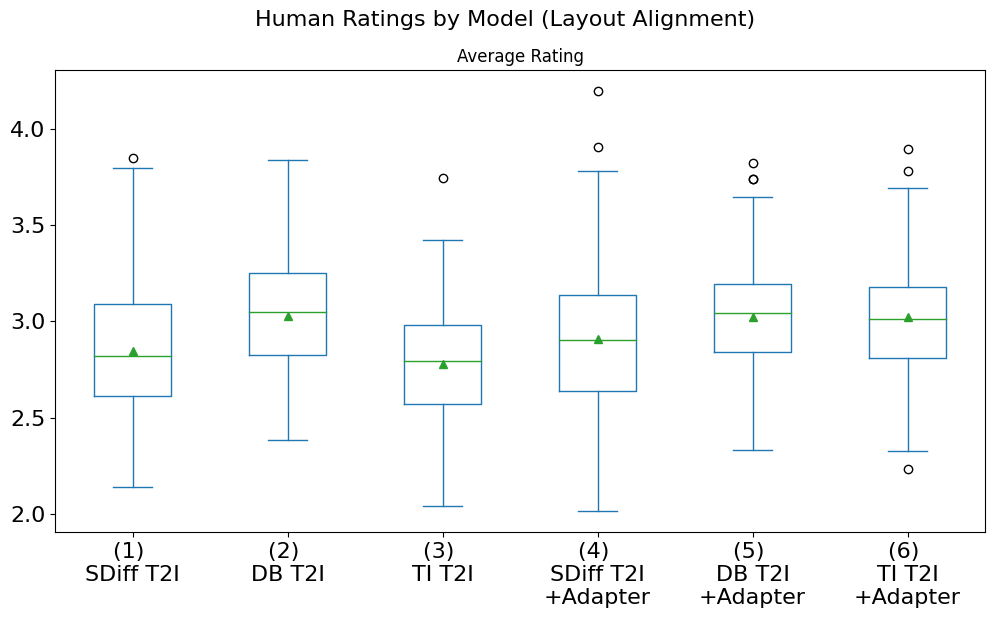}
        \caption{Layout alignment user study results.}
        \label{fig:boxplot-layout}
    \end{subfigure}
    \caption{All ratings have been normalized for every user and then scaled back to range 1-5. Human ratings of images generated with different models show robust rankings for (a) image fidelity and (b) text alignment. Results for (c) layout alignment seem less expressive, although slight differences are still visible. Humans rate real images consistently as most realistic, but there are substantial differences between models w.r.t. fidelity and alignment scores.}
    \label{fig:boxplots}
\end{figure*}

\begin{figure*}[ht!]
    \begin{subfigure}{0.33\textwidth}
        \centering
        \includegraphics[width=\textwidth]{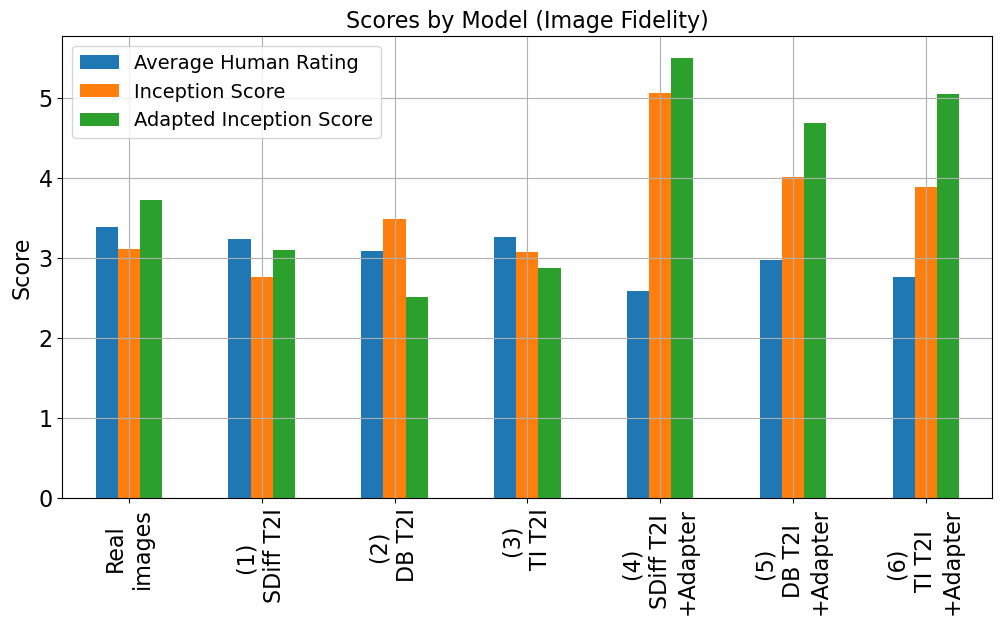}
        \caption{Image fidelity all results.}
        \label{fig:barplot-fidelity}
    \end{subfigure}
    \begin{subfigure}{0.33\textwidth}
        \centering
        \includegraphics[width=\textwidth]{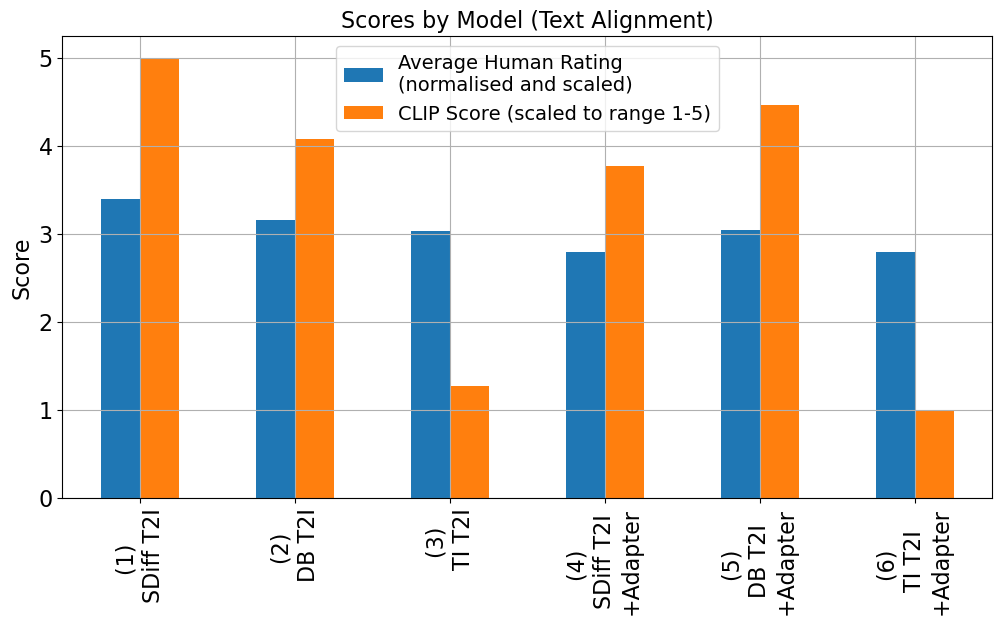}
        \caption{Text alignment all results.}
        \label{fig:barplot-text}
    \end{subfigure}
    \begin{subfigure}{0.33\textwidth}
        \centering
        \includegraphics[width=\textwidth]{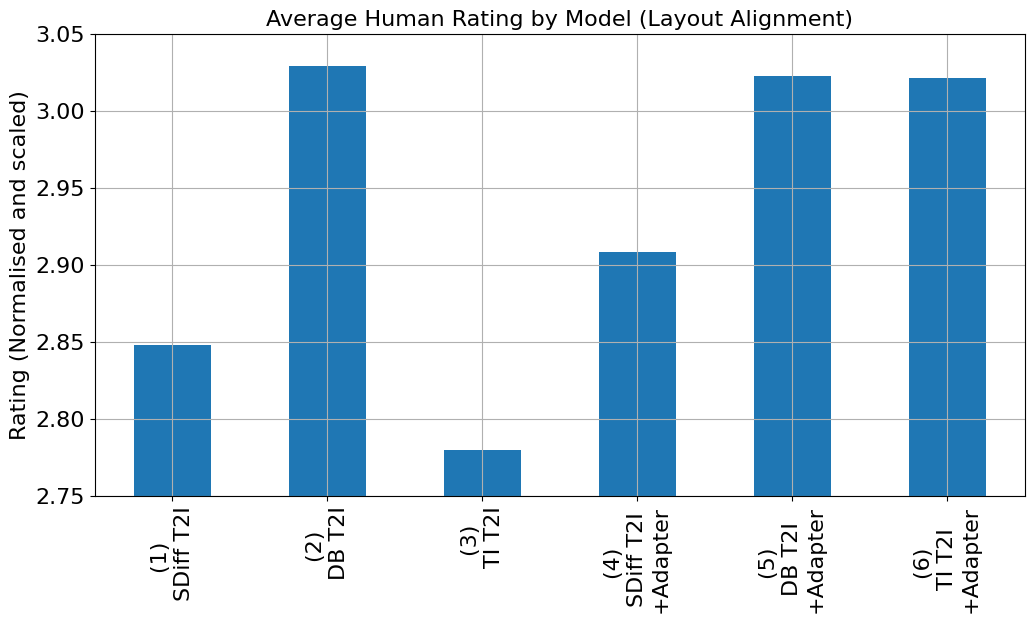}
        \caption{Layout alignment all results.}
        \label{fig:barplot-layout}
    \end{subfigure}
    \caption{The plots show the quality assessment of images generated with different models. Results are depicted by metric for each model for experiments based on (a) image fidelity, (b) text alignment and (c) layout alignment. A comparison of human ratings and offline metrics demonstrates that there's a disparity between the two: Image quality of different models is judged differently, images getting a high rating from humans can get a low score from offline metrics.}
    \label{fig:barplots}
    \vspace{-1.0em}
\end{figure*}

In regard to the compatibility of image-text pairs, a common metric is the CLIPScore \cite{hesselclipscore2021}, which relies on the CLIP \cite{radford_clip_2021} model: It measures the image-text similarity, thus the higher the score, the better. The CLIPScore is defined by the cosine similarity between the image CLIP embeddings $\boldsymbol{c}$ and text CLIP embeddings $\boldsymbol{v}$ (see \autoref{eq:clipscore}). The score is bound between 0 and 100.
\begin{equation} \label{eq:clipscore}
  \text{CLIPScore}(\boldsymbol{c},\boldsymbol{v}) =  \max(100*\cos{(\boldsymbol{c},\boldsymbol{v})}, 0)
\end{equation} 
For our experiments, CLIPScore is calculated using an open-source package \cite{taited2023CLIPScore}, the default CLIP model is ViT-B/32.

For correlation analysis, we investigate the relationship between the automated metrics and the respective human ratings. For this, we calculate the standard Pearson and Kendall Tau correlation coefficients, and the Spearman rank correlation. 

Regarding the obtained human ratings, we have to consider that users rate differently than others and might in general be more generous or pessimistic with their judgements. Therefore, we normalize the scores to reduce the individual user bias and spread: From each available rating $x$ of a user $j$, we subtract the mean of the ratings of that user, $\bar{x}_j$, to obtain the normalized value $\tilde{x}_{i,j} = x_{i,j} - \bar{x}_j$. 
We then scale all ratings back to the original scale of 1-5. For each image, we then calculate the mean rating and aggregate these for each model to obtain an average score for each approach.

\section{Results}

In the following, we analyze the results of the user study and compare the human ratings with established metrics. The size of our sample after applying the quality control on the collected study results, as described in Section \ref{sec:method}, is listed in \autoref{tab:num_ratings}. For the (c) layout alignment study, one group of images had to be removed for each model due to technical error, leaving us with a sample size of 496 images for each approach.

\begin{table}[bht]
    \centering
    \caption{Number of ratings, images and participants in user study. (*From 744 image groups of four.)}
    \label{tab:num_ratings}
    \resizebox{\columnwidth}{!}{
    \begin{tabular}{lrrr}
    \hline
         & \# ratings & \# images & \# participants \\
         \hline
   (a) fidelity  &  16125& 3202 &447  \\
   (b) text alignment     & 14625 & 3000 &451  \\
   (c) layout alignment & 3412 & 2976* & 290 \\
   \hline
    \end{tabular}
    }
\end{table}

\paragraph{Image fidelity.} Regarding (a), scores are shown in \autoref{table:image-fidelity-results}. Despite the smaller sample size, we can infer that the real images achieve the best results during the human experiments, followed by the text-only approaches and then the methods combined with the additional image input. A visual depiction is shown in \autoref{fig:boxplot-fidelity}.
Excluding the real imagery, for the synthetic data the fine-tuned methods yield mostly better results than the corresponding original Stable Diffusion approaches, apart from the (2) \textit{DB T2I} method. 

\begin{table}[bht]
\centering
\caption[Image fidelity results.]{Image fidelity results. The real images achieve the best results in human evaluation but receive only average scores with the automated metrics. (*Note the lower sample size compared to the other approaches.)}
\label{table:image-fidelity-results}
\resizebox{\columnwidth}{!}{
\begin{tabular}{ l r r r r r  }
 \hline
 Model/Data & Sample Size & IS $\uparrow$ & ${\text{IS}\textsubscript{adapt.}}$ $\uparrow$ & \multicolumn{2}{c}{Human Perception $\uparrow$} \\
   &  &  &  & not normalized & normalized \\
 \hline
 Real images & 202* & 3.09±0.31 & 3.74±0.42 & \textbf{3.75±0.57} & \textbf{3.39±0.29} \\
 (1) SDiff T2I & 500 & 2.76±0.19 & 3.10±0.29 & 3.45±0.69 & 3.23±0.37 \\
 (2) DB T2I & 500 & 3.48±0.29 & 2.49±0.24
 & 3.21±0.61 & 3.09±0.30 \\
 (3) TI T2I & 500 & 3.04±0.20 & 2.85±0.28 & 3.50±0.65 & 3.26±0.30 \\
 (4) SDiff T2I+Adapter & 500 & \textbf{5.05±0.40} & \textbf{5.51±0.48} & 2.29±0.69 & 2.58±0.38 \\
 (5) DB T2I+Adapter & 500 & 3.98±0.28 & 4.72±0.30 & 3.02±0.70 & 2.98±0.34 \\
 (6) TI T2I+Adapter & 500 & 3.91±0.24 & 5.03±0.55 & 2.29±0.75 & 2.76±0.39 \\
 \hline
\end{tabular}
}
\end{table}

Using additional input with the T2I-Adapter component gives us more control over the image composition during the generation process, however, the generated images seem to lack image quality: They achieve poorer results than the pure text-based approaches (see \autoref{fig:boxplot-fidelity}). But the fine-tuned approaches (5, 6) yield better results in combination with the layout control in comparison to the original (4) base model. With the text-only approaches, the unmodified model (1) achieves results comparable to the fine-tuned models, but these images often don’t show the desired satellite perspective: This aspect is not considered with the Likert scale and was not an influencing factor for the users regarding image fidelity, however this is a limiting factor for the generation of satellite imagery. There was a significant difference between the ratings depending on what was used as input modality (e.g. canny input seems to produce lower human ratings than sketch or depth maps), however this was not further investigated in the scope of this work.

In contrast to the human ratings,  automated metrics consistently rank \textit{Adapter}-approaches higher than generative models based solely on text input (see \autoref{fig:barplot-fidelity}). Furthermore, the real images achieve a relatively average IS and ${\text{IS}\textsubscript{adapt.}}$ in comparison to the other models.

\begin{table}[bht]
\centering
\caption[Text alignment results.]{For the text alignment results, CLIPScore seems to align with human judgement. (1) achieves the best results in this study.}
\label{table:text-alignment-results}
\resizebox{\columnwidth}{!}{
\begin{tabular}{ l r r r r  }
 \hline
 Model & Sample Size & CLIPScore $\uparrow$ & \multicolumn{2}{c}{Human Perception $\uparrow$} \\
   &  &  & not normalized & normalized \\
 \hline
 (1) SDiff T2I & 500 & \textbf{32.74} & \textbf{3.62±0.69} & \textbf{3.40±0.38} \\
 (2) DB T2I & 500 & 31.24 & 3.28±0.76 & 3.17±0.39 \\
 (3) TI T2I & 500 & 26.64 & 3.00±0.83 & 3.03±0.43 \\
 (4) SDiff T2I+Adapter & 500 & 30.74 & 2.59±0.81 & 2.79±0.43 \\
 (5) DB T2I+Adapter & 500 & 31.87 & 3.05±0.75 & 3.05±0.38 \\
 (6) TI T2I+Adapter & 500 & 26.19 & 2.57±0.81 & 2.79±0.41 \\
 \hline
\end{tabular}
}
\end{table}

\paragraph{Text alignment.} Evaluation results for (b) text alignment are shown in \autoref{table:text-alignment-results}. Here, the original model (1) achieves the best image-text alignment scores from human perspective as well as the CLIPScore. Following, the DreamBooth fine-tuned approaches (2, 5) yield the second-best results. Apart from this, the text alignment seems to result in mostly poorer results with the addition of image input, according to our human evaluation, as shown in \autoref{fig:boxplot-text}. However, this ranking does not exactly align with the CLIPScore results. Here, both Textual Inversion fine-tuned approaches (3, 6) significantly perform the poorest (see \autoref{fig:barplot-text}).

\paragraph{Layout alignment.} Since there are no suitable quantitative metrics to evaluate the layout alignment across several images, at least when the viewpoint and conditions are different in each, we only look at the human judgement results for the (c) layout alignment experiments. The results are shown in \autoref{table:layout-alignment-results}: Contrary to expectations, the DreamBooth fine-tuned approach (2) without additional image input achieves the best results in our human evaluation, as also visible in \autoref{fig:barplot-layout} and \autoref{fig:boxplot-layout}. However, only by a small margin. One possible reason could be, that the raters might still have been influenced by the image quality or other distortions and details, instead of solely focusing on the layout aspect. In general, the additional conditioning input through the adapter does lead to a better structural alignment according to the scores. Except for (2), all \textit{Adapter}-approaches (4, 5, 6) outperform the ones that are solely based on text input, (1, 3). For the \textit{Adapter}-approaches, fine-tuning seems to help the retaining of the given layout structure during the generation process, as these (5, 6) achieve noticeable better scores than (4). 

\begin{table}[bht]
\centering
\caption[Layout alignment results.]{For layout alignment, only human evaluation is available. (2) performs the best, but apart from this, the \textit{Adapter}-approaches are mostly better at structural control.}
\label{table:layout-alignment-results}
\resizebox{\columnwidth}{!}{
\begin{tabular}{ l r r r  }
 \hline
 Model & Sample Size & \multicolumn{2}{c}{Human Perception $\uparrow$} \\
   &  & not normalized & normalized \\
 \hline
 (1) SDiff T2I & 496 & 2.656±0.75 & 2.848±0.36 \\
 (2) DB T2I & 496 & \textbf{2.914±0.67} & \textbf{3.029±0.30} \\
 (3) TI T2I & 496 & 2.416±0.65 & 2.780±0.29 \\
 (4) SDiff T2I+Adapter & 496 & 2.688±0.77 & 2.909±0.37 \\
 (5) DB T2I+Adapter & 496 & 2.858±0.61 & 3.023±0.29 \\
 (6) TI T2I+Adapter & 496 & 2.843±0.62 & 3.022±0.31 \\
 \hline
\end{tabular}
}
\end{table}

\begin{figure}[bht]
    \begin{subfigure}{0.5\columnwidth}
        \centering
        \includegraphics[width=\columnwidth]{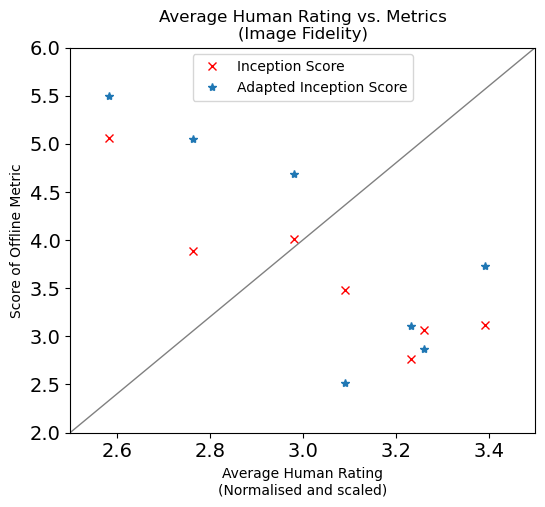}
        \captionsetup{labelformat=empty}
        \caption{(a) Image fidelity.}
        \label{fig:correlation-is}
    \end{subfigure}
    \begin{subfigure}{0.49\columnwidth}
        \centering
        \includegraphics[width=\columnwidth]{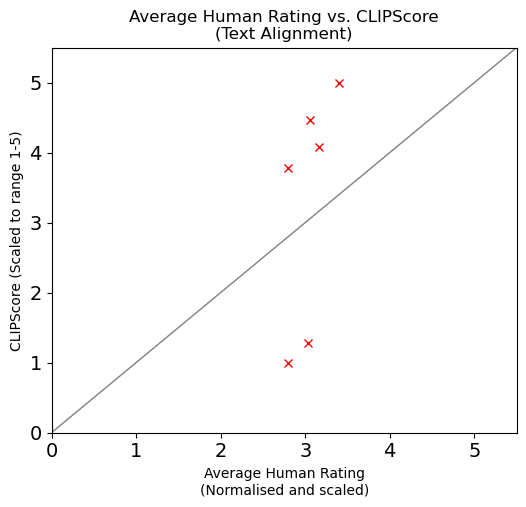}
        \captionsetup{labelformat=empty}
        \caption{(b) Text alignment.}
        \label{fig:correlation-clipscore}
    \end{subfigure}
    \caption{Visual comparison between the automatic metrics and the human ratings (normalized and scaled) from the user studies. Comparison between (a) IS scores and human judgement regarding image fidelity and (b) CLIPScore and human judgement regarding image-text alignment. There is a negative correlation visible between the IS/${\text{IS}\textsubscript{adapt.}}$ and human ratings (a). The CLIPScores and human ratings regarding text alignment correlate positively (b).}
    \label{fig:correlation-plots}
    \vspace{-1.0em}
\end{figure}

\paragraph{Correlation.} A correlation analysis has been performed between the quantitative metrics and the respective user study results, the scores are shown in \autoref{table:correlation}. For each model, the scores are visually depicted in \autoref{fig:correlation-plots}. 

For (a) image fidelity, we see that current evaluation metrics, such as the IS and also its adapted version, don’t align with human visual perception. They even correlate negatively, albeit a little less for ${\text{IS}\textsubscript{adapt.}}$ compared to the original IS, when looking at the correlation coefficients in \autoref{table:correlation}. The negative correlation is also visible in \autoref{fig:correlation-is}.
For (b) text alignment, CLIPScore seems to correlate mostly well with the human ratings, as evident by the positive scores in \autoref{table:correlation} and visually in \autoref{fig:correlation-clipscore}. However, for some models the ranking does not match that of the user studies (see \autoref{fig:barplot-text}). Furthermore, the (4) \textit{SDiff+Adapter} model achieves a relatively high score as well, although scoring second lowest according to human judgement.

\begin{table}[bht]
\centering
\caption[Correlation.]{Correlation between human judgement and automated metrics. For (a) fidelity, the metrics correlate negatively with the collected ratings. Regarding (b) text alignment, CLIPScore seems to approximate human judgement well.}
\label{table:correlation}
\resizebox{\columnwidth}{!}{
\begin{tabular}{ l l r r r  }
 \hline
  & Correlation & Pearson & Spearman & Kendall \\
 \hline
 (a) & Human Rating vs IS & -0.91 & -0.82 & -0.62 \\
  & Human Rating vs ${\text{IS}\textsubscript{adapt.}}$ & -0.79 & -0.68 & -0.52 \\
 \hline
 (b) & Human Rating vs CLIPScore & 0.61 & 0.89 & 0.73 \\
 \hline
\end{tabular}
}
\end{table}

\section{Discussion}

In this work we investigated to what extent modern generative DL methods can be used to generate imagery of rare objects in niche domains. A special focus in this work was on a comparison of control mechanisms for generative methods. In order to compare different approaches, we leveraged automated quantitative metrics and compared them with human ratings. In extensive empirical evaluations, we demonstrate that novel image generation methods can be used to generate imagery from niche domains and rare objects. Importantly, we find that textual control works well in many cases. But also controlling the image generation with building layouts is feasible, which allows for more fine-grained control.

\paragraph{Inception Score is not aligned with Human Ratings.} A key finding of our empirical results, which is in line with previous studies \cite{Kolchinski_2019}, is that the automated metrics that are used to optimize and evaluate image generation for rare object classes, do not capture image quality as rated by humans.
Our results show that exchanging the off-the-shelf classifier models pretrained on ImageNet, which are used as feature extraction backbone in current metrics like the IS, might provide a slightly better metric. However, our results also demonstrate that despite these adaptations to the domain of interest, the IS is not aligned with human visual perception. According to \cite{stein2023exposing}, established metrics as used with the default Inception model as backbone might even behave unfair towards diffusion models. Thus, current metrics, as is, are not a reliable measure of performance and image quality when the benchmark is human perception and the goal of these metrics is to actually approximate human judgement. Especially for rare classes and niche domains, as in our case, established metrics are not a trustworthy method of evaluation. Since in such cases, there is not enough real data to calculate additional comparative metrics (such as the FID, KID, Precision and Recall) to get a more robust and broader spectrum of evaluation, the main current established metric for fidelity is the IS, which, as seen in our experiments, is not reliable. Surprisingly, the IS even correlates negatively with human judgement. This appears to suggest a meaningful relationship between these two aspects -- but in the opposite direction from what the IS score is intended to measure. 

\paragraph{CLIPScore and Human Ratings.} As seen in our (b) text alignment results, CLIPScore seems to correlate mostly well with human judgement, there might only be a slight bias towards the (1) original base model, e.g. (4) \textit{SDiff+Adapter} gets a higher CLIPScore while scoring low in the human experiments. This could be due to both the model and the metric relying on CLIP (Stable Diffusion v1.5 uses the pretrained text encoder CLIP ViT-L/14 as conditioning component). The textual inversion fine-tuned approaches, for example, perform the poorest according to CLIPScore. Fine-tuning with this method trains in the textual embedding space, thus makes changes to the text encoder component, which would reinforce the assumption. Note that while the positive correlation of CLIPScore suggests that this reliably captures human perception, there could be other explanations for this positive result that we cannot rule out based on our user studies alone. For instance, if the generated images always showed the same (or very similar) facilities, this could lead to high scores in our user studies. Heterogeneity of facilities was not enforced in those models that did not use layout inputs. As both the generating models as well as the score use CLIP backbones, such cases could be regarded as a case of overfitting. This problem has been widely acknowledged: the more expressive the models, the more difficult benchmarking becomes as it can be difficult to ensure a clean train-test split \cite{srivastavaImitationGameQuantifying2023}.

For (c) structural alignment, the approaches using additional image input specifying the precise layout of a facility would, mostly outperform the ones using only text conditioning. This holds true for most of the methods except for (2) \textit{DB T2I}, which scores highest in terms of alignment -- but uses only textual input and no layout input (see \autoref{table:layout-alignment-results}). The reasons for this are unclear, one explanation could be that the user study design was not adequate enough to measure the structural control, or the instructions were not clear enough. Since we only have the human ratings to interpret, comparative analysis to other metrics is not possible, thus, a broad and robust evaluation is difficult. The layout alignment - with still considering different rotations and angles from the same target object - is an aspect which has not been thoroughly evaluated in literature yet and could be investigated in further research.

Looking at the evaluation of generative models and their outputs, the question is also raised whether human judgement should be the standard for assessing image quality: Human perception should be used as benchmark when the goal is to approximate this through a metric. However, approximating human perception is not necessarily relevant for all use cases, as also raised in other work \cite{stein2023exposing}. In fact, some of the most interesting application scenarios, such as generating new training data for ML models, do not involve human perception directly.

\paragraph{Societal implications} In the context of public interest, human-centric approaches and human-in-the-loop methods are important to understand the risks of ML technology \cite{shi_socialgood_2020, Floridi2020}. For instance, previous work in computer vision has demonstrated that it is possible to decouple human perception from machine perception with synthetic imagery \cite{athalyeSynthesizingRobustAdversarial2018}. Complementing this work on divergent perception between humans and machines, our findings show a misalignment between human visual perception and automated evaluation, which provides a flawed foundation and benchmark for future development of novel ML technologies.  
Albeit the participants in our studies are no experts in the domain, they are representatives of the general public. 
Given fabricated or generated imagery, spreading misinformation is even easier when people are not familiar with such niche content. Works addressing information manipulation via ML and crowdsourcing have gained traction within the ``AI for Social Good'' community \cite{shi_socialgood_2020}, possibly due to the rise of fake news, deepfakes and their effortless distribution through mass media. With easily available technology, virtually anyone can produce synthetic imagery that can, as shown in our experiments, fool the average user. Systems being open to validation by, e.g. being open-source, is necessary for transparency \cite{zueger2022}. However, an ill-intended user could use such powerful open-source technology for malicious purposes. In our studies, generated images look authentic enough for users to assume they are real, which, depending on the content, could have implications of public interest for citizens.

\section{Conclusion}

In this work, we have leveraged a pre-trained vision-language model and fine-tuned it to generate synthetic satellite imagery of a rare object class, which has been underrepresented in literature before. Moreover, we conducted large-scale human-in-the-loop experiments to measure human judgement and compared it with established metrics in the field. We found that additional image input mostly gives more control over the image composition, however, it still remains very difficult to control specific details and generate images of the same exact object with the presented conditioning methods. Our results demonstrate that fine-tuning can help generate imagery of specific images and target objects that are on par with data generated from the original base model, in terms of perceived image quality, but that are more suitable for the remote sensing domain and better display the desired satellite perspective. Consistent with previous works, we confirm that established state-of-the-art metrics to evaluate synthetic imagery do not necessarily align with human perception, at least regarding image fidelity. Our findings show that the IS and its adapted version even correlate negatively. CLIPScore seems to work fairly well for measuring image-text alignment, but might be biased towards models based on CLIP. Overall, we find that large-scale user studies are needed to assess synthetic data in regard to human perception, especially for rare classes, where a broad variety of automated evaluation metrics is not available.

For future work, these experiments could be conducted on an even larger scale and for various datasets also in other domains, to investigate whether the findings of this work generalize to other use cases. In line with previous work, our results provide empirical evidence that current established metrics do not work well for measuring human judgement, especially for rare objects and domains that contain imagery dissimilar to natural images. The quantitative evaluation with automated metrics in genAI requires a more in-depth study and remains an open field of research, not only for the sake of evaluation itself: Better understanding of the perceived image quality will enable researchers to improve generative models in the future.

\begin{acknowledgments}

This research is funded by the German Foundation for Peace Research (Deutsche Stiftung Friedensforschung) and is part of the project ``Citizen-Based Monitoring for Peace \& Security in the Era of Synthetic Media and Deepfakes'' (FP 06/22 | 01/21-
FB3-AdD-Pro). Felix Bießmann received funding from the Einstein Center Digital Future, Berlin and the German Research Foundation (DFG) - Project number: 528483508 - FIP 12.

\end{acknowledgments}

\bibliography{ceur}

\begin{thebibliography}{57}
\expandafter\ifx\csname natexlab\endcsname\relax\def\natexlab#1{#1}\fi
\providecommand{\url}[1]{\texttt{#1}}
\providecommand{\href}[2]{#2}
\providecommand{\path}[1]{#1}
\providecommand{\DOIprefix}{doi:}
\providecommand{\ArXivprefix}{arXiv:}
\providecommand{\URLprefix}{URL: }
\providecommand{\Pubmedprefix}{pmid:}
\providecommand{\doi}[1]{\href{http://dx.doi.org/#1}{\path{#1}}}
\providecommand{\Pubmed}[1]{\href{pmid:#1}{\path{#1}}}
\providecommand{\bibinfo}[2]{#2}
\ifx\xfnm\relax \def\xfnm[#1]{\unskip,\space#1}\fi
\bibitem[{Fonseca and Bacao(2023)}]{fonsecaTabularLatentSpace2023a}
\bibinfo{author}{J.~Fonseca}, \bibinfo{author}{F.~Bacao},
\newblock \bibinfo{title}{Tabular and latent space synthetic data generation: a
  literature review},
\newblock \bibinfo{journal}{Journal of Big Data} \bibinfo{volume}{10}
  (\bibinfo{year}{2023}) \bibinfo{pages}{115}.
\bibitem[{Hassan et~al.(2023)Hassan, Salomone, and
  Mengersen}]{hassanDeepGenerativeModels2023}
\bibinfo{author}{C.~Hassan}, \bibinfo{author}{R.~Salomone},
  \bibinfo{author}{K.~Mengersen}, \bibinfo{title}{Deep {Generative} {Models},
  {Synthetic} {Tabular} {Data}, and {Differential} {Privacy}: {An} {Overview}
  and {Synthesis}}, \bibinfo{year}{2023}. \bibinfo{note}{ArXiv:2307.15424 [cs,
  stat]}.
\bibitem[{Zhao et~al.(2021)Zhao, Zhang, Xu, Sun, and Deng}]{Zhao2021}
\bibinfo{author}{B.~Zhao}, \bibinfo{author}{S.~Zhang}, \bibinfo{author}{C.~Xu},
  \bibinfo{author}{Y.~Sun}, \bibinfo{author}{C.~Deng},
\newblock \bibinfo{title}{Deep fake geography? when geospatial data encounter
  artificial intelligence},
\newblock \bibinfo{journal}{Cartography and Geographic Information Science}
  \bibinfo{volume}{48} (\bibinfo{year}{2021}) \bibinfo{pages}{338--352}.
\bibitem[{Rombach et~al.(2022)Rombach, Blattmann, Lorenz, Esser, and
  Ommer}]{rombachhighresolution2022}
\bibinfo{author}{R.~Rombach}, \bibinfo{author}{A.~Blattmann},
  \bibinfo{author}{D.~Lorenz}, \bibinfo{author}{P.~Esser},
  \bibinfo{author}{B.~Ommer},
\newblock \bibinfo{title}{{High-Resolution Image Synthesis with Latent
  Diffusion Models}},
\newblock in: \bibinfo{booktitle}{{IEEE/CVF} Conference on Computer Vision and
  Pattern Recognition, {CVPR} 2022, New Orleans, LA, USA, June 18-24, 2022},
  \bibinfo{publisher}{{IEEE}}, \bibinfo{year}{2022}, pp.
  \bibinfo{pages}{10674--10685}.
\bibitem[{{Center for Security and Emerging Technology} et~al.(2021){Center for
  Security and Emerging Technology}, Buchanan, Lohn, Musser, and
  Sedova}]{centerforsecurityandemergingtechnologyTruthLiesAutomation2021}
\bibinfo{author}{{Center for Security and Emerging Technology}},
  \bibinfo{author}{B.~Buchanan}, \bibinfo{author}{A.~Lohn},
  \bibinfo{author}{M.~Musser}, \bibinfo{author}{K.~Sedova},
  \bibinfo{title}{Truth, {Lies}, and {Automation}: {How} {Language} {Models}
  {Could} {Change} {Disinformation}}, \bibinfo{type}{Technical Report}, Center
  for Security and Emerging Technology, \bibinfo{year}{2021}. \URLprefix
  \url{https://cset.georgetown.edu/publication/truth-lies-and-automation/}.
  \DOIprefix\doi{10.51593/2021CA003}.
\bibitem[{Metz et~al.(2024)Metz, Kang, Frenkel, Thompson, and
  Grant}]{cademetzHowTechGiants}
\bibinfo{author}{C.~Metz}, \bibinfo{author}{C.~Kang},
  \bibinfo{author}{S.~Frenkel}, \bibinfo{author}{S.~A. Thompson},
  \bibinfo{author}{N.~Grant},
\newblock \bibinfo{title}{How {Tech} {Giants} {Cut} {Corners} to {Harvest}
  {Data} for {A}.{I}.},
\newblock \bibinfo{journal}{The New York Times}  (\bibinfo{year}{2024}).
  \URLprefix
  \url{https://www.nytimes.com/2024/04/06/technology/tech-giants-harvest-data-artificial-intelligence.html}.
\bibitem[{Chen et~al.(2021)Chen, Lu, Chen, Williamson, and
  Mahmood}]{chenSyntheticDataMachine2021}
\bibinfo{author}{R.~J. Chen}, \bibinfo{author}{M.~Y. Lu},
  \bibinfo{author}{T.~Y. Chen}, \bibinfo{author}{D.~F.~K. Williamson},
  \bibinfo{author}{F.~Mahmood},
\newblock \bibinfo{title}{Synthetic data in machine learning for medicine and
  healthcare},
\newblock \bibinfo{journal}{Nature Biomedical Engineering} \bibinfo{volume}{5}
  (\bibinfo{year}{2021}) \bibinfo{pages}{493--497}.
\bibitem[{Zhou et~al.(2019)Zhou, Gordon, Krishna, Narcomey, Fei-Fei, and
  Bernstein}]{zhou2019hype}
\bibinfo{author}{S.~Zhou}, \bibinfo{author}{M.~L. Gordon},
  \bibinfo{author}{R.~Krishna}, \bibinfo{author}{A.~Narcomey},
  \bibinfo{author}{L.~Fei-Fei}, \bibinfo{author}{M.~S. Bernstein},
  \bibinfo{title}{Hype: A benchmark for human eye perceptual evaluation of
  generative models}, \bibinfo{year}{2019}.
  \href{http://arxiv.org/abs/1904.01121}{{\tt arXiv:1904.01121}}.
\bibitem[{Goodfellow et~al.(2014)Goodfellow, Pouget-Abadie, Mirza, Xu,
  Warde-Farley, Ozair, Courville, and Bengio}]{goodfellow_gan_2014}
\bibinfo{author}{I.~J. Goodfellow}, \bibinfo{author}{J.~Pouget-Abadie},
  \bibinfo{author}{M.~Mirza}, \bibinfo{author}{B.~Xu},
  \bibinfo{author}{D.~Warde-Farley}, \bibinfo{author}{S.~Ozair},
  \bibinfo{author}{A.~Courville}, \bibinfo{author}{Y.~Bengio},
  \bibinfo{title}{Generative adversarial networks}, \bibinfo{year}{2014}.
\bibitem[{Shamsolmoali et~al.(2021)Shamsolmoali, Zareapoor, Granger, Zhou,
  Wang, Celebi, and Yang}]{shamsolmoaliimage2020}
\bibinfo{author}{P.~Shamsolmoali}, \bibinfo{author}{M.~Zareapoor},
  \bibinfo{author}{E.~Granger}, \bibinfo{author}{H.~Zhou},
  \bibinfo{author}{R.~Wang}, \bibinfo{author}{M.~E. Celebi},
  \bibinfo{author}{J.~Yang},
\newblock \bibinfo{title}{{Image Synthesis with Adversarial Networks: a
  Comprehensive Survey and Case Studies}},
\newblock \bibinfo{journal}{Inf. Fusion} \bibinfo{volume}{72}
  (\bibinfo{year}{2021}) \bibinfo{pages}{126--146}.
\bibitem[{Ho et~al.(2020)Ho, Jain, and Abbeel}]{ho_denoising_2020}
\bibinfo{author}{J.~Ho}, \bibinfo{author}{A.~Jain},
  \bibinfo{author}{P.~Abbeel}, \bibinfo{title}{Denoising {Diffusion}
  {Probabilistic} {Models}}, \bibinfo{year}{2020}.
  \bibinfo{note}{ArXiv:2006.11239 [cs, stat]}.
\bibitem[{Dhariwal and Nichol(2021)}]{dhariwaldiffusion2021}
\bibinfo{author}{P.~Dhariwal}, \bibinfo{author}{A.~Q. Nichol},
\newblock \bibinfo{title}{{Diffusion Models Beat GANs on Image Synthesis}},
\newblock in: \bibinfo{editor}{M.~Ranzato}, \bibinfo{editor}{A.~Beygelzimer},
  \bibinfo{editor}{Y.~N. Dauphin}, \bibinfo{editor}{P.~Liang},
  \bibinfo{editor}{J.~W. Vaughan} (Eds.), \bibinfo{booktitle}{Advances in
  Neural Information Processing Systems 34, NeurIPS 2021, virtual},
  \bibinfo{year}{2021}, pp. \bibinfo{pages}{8780--8794}.
\bibitem[{Nichol et~al.(2022)Nichol, Dhariwal, Ramesh, Shyam, Mishkin, McGrew,
  Sutskever, and Chen}]{nicholglide2022}
\bibinfo{author}{A.~Q. Nichol}, \bibinfo{author}{P.~Dhariwal},
  \bibinfo{author}{A.~Ramesh}, \bibinfo{author}{P.~Shyam},
  \bibinfo{author}{P.~Mishkin}, \bibinfo{author}{B.~McGrew},
  \bibinfo{author}{I.~Sutskever}, \bibinfo{author}{M.~Chen},
\newblock \bibinfo{title}{{GLIDE: Towards Photorealistic Image Generation and
  Editing with Text-Guided Diffusion Models}},
\newblock in: \bibinfo{booktitle}{International Conference on Machine Learning,
  {ICML} 2022, 17-23 July 2022, Baltimore, Maryland, {USA}}, volume
  \bibinfo{volume}{162} of \textit{\bibinfo{series}{Proceedings of Machine
  Learning Research}}, \bibinfo{publisher}{{PMLR}}, \bibinfo{year}{2022}, pp.
  \bibinfo{pages}{16784--16804}.
\bibitem[{Ramesh et~al.(2022)Ramesh, Dhariwal, Nichol, Chu, and
  Chen}]{ramesh2022hierarchical}
\bibinfo{author}{A.~Ramesh}, \bibinfo{author}{P.~Dhariwal},
  \bibinfo{author}{A.~Nichol}, \bibinfo{author}{C.~Chu},
  \bibinfo{author}{M.~Chen}, \bibinfo{title}{Hierarchical text-conditional
  image generation with clip latents}, \bibinfo{year}{2022}.
  \href{http://arxiv.org/abs/2204.06125}{{\tt arXiv:2204.06125}}.
\bibitem[{Ramesh et~al.(2021)Ramesh, Pavlov, Goh, Gray, Voss, Radford, Chen,
  and Sutskever}]{rameshzeroshot2021}
\bibinfo{author}{A.~Ramesh}, \bibinfo{author}{M.~Pavlov},
  \bibinfo{author}{G.~Goh}, \bibinfo{author}{S.~Gray},
  \bibinfo{author}{C.~Voss}, \bibinfo{author}{A.~Radford},
  \bibinfo{author}{M.~Chen}, \bibinfo{author}{I.~Sutskever},
\newblock \bibinfo{title}{{Zero-Shot Text-to-Image Generation}},
\newblock in: \bibinfo{editor}{M.~Meila}, \bibinfo{editor}{T.~Zhang} (Eds.),
  \bibinfo{booktitle}{Proceedings of the 38th International Conference on
  Machine Learning, {ICML} 2021, Virtual Event}, volume \bibinfo{volume}{139}
  of \textit{\bibinfo{series}{Proceedings of Machine Learning Research}},
  \bibinfo{publisher}{{PMLR}}, \bibinfo{year}{2021}, pp.
  \bibinfo{pages}{8821--8831}.
\bibitem[{Saharia et~al.(2022)Saharia, Chan, Saxena, Li, Whang, Denton,
  Ghasemipour, Lopes, Ayan, Salimans, Ho, Fleet, and
  Norouzi}]{sahariaphotorealistic2022}
\bibinfo{author}{C.~Saharia}, \bibinfo{author}{W.~Chan},
  \bibinfo{author}{S.~Saxena}, \bibinfo{author}{L.~Li},
  \bibinfo{author}{J.~Whang}, \bibinfo{author}{E.~L. Denton},
  \bibinfo{author}{S.~K.~S. Ghasemipour}, \bibinfo{author}{R.~G. Lopes},
  \bibinfo{author}{B.~K. Ayan}, \bibinfo{author}{T.~Salimans},
  \bibinfo{author}{J.~Ho}, \bibinfo{author}{D.~J. Fleet},
  \bibinfo{author}{M.~Norouzi},
\newblock \bibinfo{title}{{Photorealistic Text-to-Image Diffusion Models with
  Deep Language Understanding}},
\newblock in: \bibinfo{booktitle}{NeurIPS}, \bibinfo{year}{2022}.
\bibitem[{Esser et~al.(2019)Esser, Haux, and Ommer}]{Esser_2019_ICCV}
\bibinfo{author}{P.~Esser}, \bibinfo{author}{J.~Haux},
  \bibinfo{author}{B.~Ommer},
\newblock \bibinfo{title}{Unsupervised robust disentangling of latent
  characteristics for image synthesis},
\newblock in: \bibinfo{booktitle}{Proceedings of the IEEE/CVF International
  Conference on Computer Vision (ICCV)}, \bibinfo{year}{2019}.
\bibitem[{H\"{a}rk\"{o}nen et~al.(2020)H\"{a}rk\"{o}nen, Hertzmann, Lehtinen,
  and Paris}]{NEURIPS2020_6fe43269}
\bibinfo{author}{E.~H\"{a}rk\"{o}nen}, \bibinfo{author}{A.~Hertzmann},
  \bibinfo{author}{J.~Lehtinen}, \bibinfo{author}{S.~Paris},
\newblock \bibinfo{title}{Ganspace: Discovering interpretable gan controls},
\newblock in: \bibinfo{editor}{H.~Larochelle}, \bibinfo{editor}{M.~Ranzato},
  \bibinfo{editor}{R.~Hadsell}, \bibinfo{editor}{M.~Balcan},
  \bibinfo{editor}{H.~Lin} (Eds.), \bibinfo{booktitle}{Advances in Neural
  Information Processing Systems}, volume~\bibinfo{volume}{33},
  \bibinfo{publisher}{Curran Associates, Inc.}, \bibinfo{year}{2020}, pp.
  \bibinfo{pages}{9841--9850}.
\bibitem[{Park et~al.(2023)Park, Kwon, Choi, Jo, and Uh}]{NEURIPS2023_4bfcebed}
\bibinfo{author}{Y.-H. Park}, \bibinfo{author}{M.~Kwon},
  \bibinfo{author}{J.~Choi}, \bibinfo{author}{J.~Jo}, \bibinfo{author}{Y.~Uh},
\newblock \bibinfo{title}{Understanding the latent space of diffusion models
  through the lens of riemannian geometry},
\newblock in: \bibinfo{editor}{A.~Oh}, \bibinfo{editor}{T.~Neumann},
  \bibinfo{editor}{A.~Globerson}, \bibinfo{editor}{K.~Saenko},
  \bibinfo{editor}{M.~Hardt}, \bibinfo{editor}{S.~Levine} (Eds.),
  \bibinfo{booktitle}{Advances in Neural Information Processing Systems},
  volume~\bibinfo{volume}{36}, \bibinfo{publisher}{Curran Associates, Inc.},
  \bibinfo{year}{2023}, pp. \bibinfo{pages}{24129--24142}.
\bibitem[{Ma et~al.(2019)Ma, Liu, Zhang, Ye, Yin, and Johnson}]{ma_deep_2019}
\bibinfo{author}{L.~Ma}, \bibinfo{author}{Y.~Liu}, \bibinfo{author}{X.~Zhang},
  \bibinfo{author}{Y.~Ye}, \bibinfo{author}{G.~Yin}, \bibinfo{author}{B.~A.
  Johnson},
\newblock \bibinfo{title}{Deep learning in remote sensing applications: {A}
  meta-analysis and review},
\newblock \bibinfo{journal}{ISPRS Journal of Photogrammetry and Remote Sensing}
  \bibinfo{volume}{152} (\bibinfo{year}{2019}) \bibinfo{pages}{166--177}.
\bibitem[{Xu and Zhao(2018)}]{xu_satellite_2018}
\bibinfo{author}{C.~Xu}, \bibinfo{author}{B.~Zhao},
\newblock \bibinfo{title}{{Satellite Image Spoofing: Creating Remote Sensing
  Dataset with Generative Adversarial Networks (Short Paper)}},
\newblock in: \bibinfo{editor}{S.~Winter}, \bibinfo{editor}{A.~Griffin},
  \bibinfo{editor}{M.~Sester} (Eds.), \bibinfo{booktitle}{10th International
  Conference on Geographic Information Science (GIScience 2018)}, volume
  \bibinfo{volume}{114} of \textit{\bibinfo{series}{Leibniz International
  Proceedings in Informatics (LIPIcs)}}, \bibinfo{publisher}{Schloss
  Dagstuhl--Leibniz-Zentrum fuer Informatik}, \bibinfo{address}{Dagstuhl,
  Germany}, \bibinfo{year}{2018}, pp. \bibinfo{pages}{67:1--67:6}.
\bibitem[{Yates et~al.(2022)Yates, Hart, Houghton, Torres~Torres, and
  Pound}]{yatesevaluation2022}
\bibinfo{author}{M.~Yates}, \bibinfo{author}{G.~Hart},
  \bibinfo{author}{R.~Houghton}, \bibinfo{author}{M.~Torres~Torres},
  \bibinfo{author}{M.~Pound},
\newblock \bibinfo{title}{{Evaluation of synthetic aerial imagery using
  unconditional generative adversarial networks}},
\newblock \bibinfo{journal}{ISPRS Journal of Photogrammetry and Remote Sensing}
  \bibinfo{volume}{190} (\bibinfo{year}{2022}) \bibinfo{pages}{231--251}.
\bibitem[{Shermeyer et~al.(2020)Shermeyer, Hossler, Etten, Hogan, Lewis, and
  Kim}]{RarePlanes}
\bibinfo{author}{J.~Shermeyer}, \bibinfo{author}{T.~Hossler},
  \bibinfo{author}{A.~V. Etten}, \bibinfo{author}{D.~Hogan},
  \bibinfo{author}{R.~Lewis}, \bibinfo{author}{D.~Kim},
  \bibinfo{title}{Rare{P}lanes: {S}ynthetic {D}ata {T}akes {F}light},
  \bibinfo{year}{2020}. \href{http://arxiv.org/abs/2006.02963}{{\tt
  arXiv:2006.02963}}.
\bibitem[{Kong et~al.(2020)Kong, Huang, Bradbury, and Malof}]{Synthinel-1}
\bibinfo{author}{F.~Kong}, \bibinfo{author}{B.~Huang},
  \bibinfo{author}{K.~Bradbury}, \bibinfo{author}{J.~M. Malof},
  \bibinfo{title}{The {S}ynthinel-1 dataset: a collection of high resolution
  synthetic overhead imagery for building segmentation}, \bibinfo{year}{2020}.
  \href{http://arxiv.org/abs/2001.05130}{{\tt arXiv:2001.05130}}.
\bibitem[{Bejiga et~al.(2019)Bejiga, Melgani, and
  Vascotto}]{bejigaretroremote2019}
\bibinfo{author}{M.~B. Bejiga}, \bibinfo{author}{F.~Melgani},
  \bibinfo{author}{A.~Vascotto},
\newblock \bibinfo{title}{{Retro-Remote Sensing: Generating Images From Ancient
  Texts}},
\newblock \bibinfo{journal}{{IEEE} J. Sel. Top. Appl. Earth Obs. Remote. Sens.}
  \bibinfo{volume}{12} (\bibinfo{year}{2019}) \bibinfo{pages}{950--960}.
\bibitem[{Chen et~al.(2021)Chen, Ma, Yao, Lv, Yang, Li, and
  Wan}]{chenremoteaugm2021}
\bibinfo{author}{C.~Chen}, \bibinfo{author}{H.~Ma}, \bibinfo{author}{G.~Yao},
  \bibinfo{author}{N.~Lv}, \bibinfo{author}{H.~Yang}, \bibinfo{author}{C.~Li},
  \bibinfo{author}{S.~Wan},
\newblock \bibinfo{title}{{Remote Sensing Image Augmentation Based on Text
  Description for Waterside Change Detection}},
\newblock \bibinfo{journal}{Remote. Sens.} \bibinfo{volume}{13}
  (\bibinfo{year}{2021}) \bibinfo{pages}{1894}.
\bibitem[{Zhao and Shi(2022)}]{zhaotexttoremote2022}
\bibinfo{author}{R.~Zhao}, \bibinfo{author}{Z.~Shi},
\newblock \bibinfo{title}{{Text-to-Remote-Sensing-Image Generation With
  Structured Generative Adversarial Networks}},
\newblock \bibinfo{journal}{{IEEE} Geosci. Remote. Sens. Lett.}
  \bibinfo{volume}{19} (\bibinfo{year}{2022}) \bibinfo{pages}{1--5}.
\bibitem[{Xu et~al.(2022)Xu, Yu, Ghamisi, Kopp, and
  Hochreiter}]{xutxt2imgmhn2022}
\bibinfo{author}{Y.~Xu}, \bibinfo{author}{W.~Yu}, \bibinfo{author}{P.~Ghamisi},
  \bibinfo{author}{M.~Kopp}, \bibinfo{author}{S.~Hochreiter},
\newblock \bibinfo{title}{{Txt2Img-MHN: Remote Sensing Image Generation from
  Text Using Modern Hopfield Networks}},
\newblock \bibinfo{journal}{CoRR} \bibinfo{volume}{abs/2208.04441}
  (\bibinfo{year}{2022}). \href{http://arxiv.org/abs/2208.04441}{{\tt
  arXiv:2208.04441}}.
\bibitem[{Nguyen et~al.(2023)Nguyen, Glaser, and
  Biessmann}]{nguyen2023generating}
\bibinfo{author}{T.~V. Nguyen}, \bibinfo{author}{A.~Glaser},
  \bibinfo{author}{F.~Biessmann},
\newblock \bibinfo{title}{Generating synthetic satellite imagery with
  deep-learning text-to-image models -- technical challenges and implications
  for monitoring and verification},
\newblock in: \bibinfo{booktitle}{INMM/ESARDA Joint Annual Meeting},
  \bibinfo{address}{Vienna, Austria}, \bibinfo{year}{2023}.
  \href{http://arxiv.org/abs/2404.07754}{{\tt arXiv:2404.07754}}.
\bibitem[{Hoster et~al.(2023)Hoster, Al-Sayed, Biessmann, Glaser, Hildebrand,
  Moric, and Nguyen}]{hoster2023using}
\bibinfo{author}{J.~Hoster}, \bibinfo{author}{S.~Al-Sayed},
  \bibinfo{author}{F.~Biessmann}, \bibinfo{author}{A.~Glaser},
  \bibinfo{author}{K.~Hildebrand}, \bibinfo{author}{I.~Moric},
  \bibinfo{author}{T.~V. Nguyen}, \bibinfo{title}{Using game engines and
  machine learning to create synthetic satellite imagery for a tabletop
  verification exercise}, \bibinfo{year}{2023}.
  \href{http://arxiv.org/abs/2404.11461}{{\tt arXiv:2404.11461}}.
\bibitem[{Bandi et~al.(2023)Bandi, Adapa, and Kuchi}]{bandi_generativeai_2023}
\bibinfo{author}{A.~Bandi}, \bibinfo{author}{P.~V. S.~R. Adapa},
  \bibinfo{author}{Y.~E. V. P.~K. Kuchi},
\newblock \bibinfo{title}{The power of generative ai: A review of requirements,
  models, input–output formats, evaluation metrics, and challenges},
\newblock \bibinfo{journal}{Future Internet} \bibinfo{volume}{15}
  (\bibinfo{year}{2023}).
\bibitem[{Man and Chahl(2022)}]{man_review_2022}
\bibinfo{author}{K.~Man}, \bibinfo{author}{J.~Chahl},
\newblock \bibinfo{title}{A {Review} of {Synthetic} {Image} {Data} and {Its}
  {Use} in {Computer} {Vision}},
\newblock \bibinfo{journal}{Journal of Imaging} \bibinfo{volume}{8}
  (\bibinfo{year}{2022}) \bibinfo{pages}{310}.
\bibitem[{Otani et~al.(2023)Otani, Togashi, Sawai, Ishigami, Nakashima, Rahtu,
  Heikkilä, and Satoh}]{otanitoward2023}
\bibinfo{author}{M.~Otani}, \bibinfo{author}{R.~Togashi},
  \bibinfo{author}{Y.~Sawai}, \bibinfo{author}{R.~Ishigami},
  \bibinfo{author}{Y.~Nakashima}, \bibinfo{author}{E.~Rahtu},
  \bibinfo{author}{J.~Heikkilä}, \bibinfo{author}{S.~Satoh},
\newblock \bibinfo{title}{Toward verifiable and reproducible human evaluation
  for text-to-image generation},
\newblock \bibinfo{journal}{2023 IEEE/CVF Conference on Computer Vision and
  Pattern Recognition (CVPR)}  (\bibinfo{year}{2023}).
\bibitem[{Heusel et~al.(2017)Heusel, Ramsauer, Unterthiner, Nessler, and
  Hochreiter}]{heuselgans2018}
\bibinfo{author}{M.~Heusel}, \bibinfo{author}{H.~Ramsauer},
  \bibinfo{author}{T.~Unterthiner}, \bibinfo{author}{B.~Nessler},
  \bibinfo{author}{S.~Hochreiter},
\newblock \bibinfo{title}{{GANs Trained by a Two Time-Scale Update Rule
  Converge to a Local Nash Equilibrium}},
\newblock in: \bibinfo{booktitle}{Advances in Neural Information Processing
  Systems 30: Annual Conference on Neural Information Processing Systems 2017,
  December 4-9, 2017, Long Beach, CA, {USA}}, \bibinfo{year}{2017}, pp.
  \bibinfo{pages}{6626--6637}.
\bibitem[{Salimans et~al.(2016)Salimans, Goodfellow, Zaremba, Cheung, Radford,
  and Chen}]{salimansimproved2016}
\bibinfo{author}{T.~Salimans}, \bibinfo{author}{I.~J. Goodfellow},
  \bibinfo{author}{W.~Zaremba}, \bibinfo{author}{V.~Cheung},
  \bibinfo{author}{A.~Radford}, \bibinfo{author}{X.~Chen},
\newblock \bibinfo{title}{{Improved Techniques for Training GANs}},
\newblock in: \bibinfo{booktitle}{Advances in Neural Information Processing
  Systems 29: Annual Conference on Neural Information Processing Systems 2016,
  December 5-10, 2016, Barcelona, Spain}, \bibinfo{year}{2016}, pp.
  \bibinfo{pages}{2226--2234}.
\bibitem[{Stein et~al.(2023)Stein, Cresswell, Hosseinzadeh, Sui, Ross,
  Villecroze, Liu, Caterini, Taylor, and Loaiza-Ganem}]{stein2023exposing}
\bibinfo{author}{G.~Stein}, \bibinfo{author}{J.~C. Cresswell},
  \bibinfo{author}{R.~Hosseinzadeh}, \bibinfo{author}{Y.~Sui},
  \bibinfo{author}{B.~L. Ross}, \bibinfo{author}{V.~Villecroze},
  \bibinfo{author}{Z.~Liu}, \bibinfo{author}{A.~L. Caterini},
  \bibinfo{author}{J.~E.~T. Taylor}, \bibinfo{author}{G.~Loaiza-Ganem},
  \bibinfo{title}{Exposing flaws of generative model evaluation metrics and
  their unfair treatment of diffusion models}, \bibinfo{year}{2023}.
  \href{http://arxiv.org/abs/2306.04675}{{\tt arXiv:2306.04675}}.
\bibitem[{Chambon et~al.(2022)Chambon, Bluethgen, Langlotz, and
  Chaudhari}]{chambon_adapting_2022}
\bibinfo{author}{P.~Chambon}, \bibinfo{author}{C.~Bluethgen},
  \bibinfo{author}{C.~P. Langlotz}, \bibinfo{author}{A.~Chaudhari},
  \bibinfo{title}{Adapting {Pretrained} {Vision}-{Language} {Foundational}
  {Models} to {Medical} {Imaging} {Domains}}, \bibinfo{year}{2022}.
  \bibinfo{note}{ArXiv:2210.04133 [cs]}.
\bibitem[{Ding et~al.(2022)Ding, Zheng, Hong, and Tang}]{ding2022cogview2}
\bibinfo{author}{M.~Ding}, \bibinfo{author}{W.~Zheng},
  \bibinfo{author}{W.~Hong}, \bibinfo{author}{J.~Tang},
  \bibinfo{title}{Cogview2: Faster and better text-to-image generation via
  hierarchical transformers}, \bibinfo{year}{2022}.
  \href{http://arxiv.org/abs/2204.14217}{{\tt arXiv:2204.14217}}.
\bibitem[{Funke et~al.(2021)Funke, Borowski, Stosio, Brendel, Wallis, and
  Bethge}]{Funke_2021}
\bibinfo{author}{C.~M. Funke}, \bibinfo{author}{J.~Borowski},
  \bibinfo{author}{K.~Stosio}, \bibinfo{author}{W.~Brendel},
  \bibinfo{author}{T.~S.~A. Wallis}, \bibinfo{author}{M.~Bethge},
\newblock \bibinfo{title}{Five points to check when comparing visual perception
  in humans and machines},
\newblock \bibinfo{journal}{Journal of Vision} \bibinfo{volume}{21}
  (\bibinfo{year}{2021}) \bibinfo{pages}{16}.
\bibitem[{Ruiz et~al.(2022)Ruiz, Li, Jampani, Pritch, Rubinstein, and
  Aberman}]{ruizdreambooth2022}
\bibinfo{author}{N.~Ruiz}, \bibinfo{author}{Y.~Li},
  \bibinfo{author}{V.~Jampani}, \bibinfo{author}{Y.~Pritch},
  \bibinfo{author}{M.~Rubinstein}, \bibinfo{author}{K.~Aberman},
\newblock \bibinfo{title}{{DreamBooth: Fine Tuning Text-to-Image Diffusion
  Models for Subject-Driven Generation}},
\newblock \bibinfo{journal}{CoRR} \bibinfo{volume}{abs/2208.12242}
  (\bibinfo{year}{2022}). \href{http://arxiv.org/abs/2208.12242}{{\tt
  arXiv:2208.12242}}.
\bibitem[{Gal et~al.(2022)Gal, Alaluf, Atzmon, Patashnik, Bermano, Chechik, and
  Cohen{-}Or}]{galimage2022}
\bibinfo{author}{R.~Gal}, \bibinfo{author}{Y.~Alaluf},
  \bibinfo{author}{Y.~Atzmon}, \bibinfo{author}{O.~Patashnik},
  \bibinfo{author}{A.~H. Bermano}, \bibinfo{author}{G.~Chechik},
  \bibinfo{author}{D.~Cohen{-}Or},
\newblock \bibinfo{title}{{An Image is Worth One Word: Personalizing
  Text-to-Image Generation using Textual Inversion}},
\newblock \bibinfo{journal}{CoRR} \bibinfo{volume}{abs/2208.01618}
  (\bibinfo{year}{2022}). \href{http://arxiv.org/abs/2208.01618}{{\tt
  arXiv:2208.01618}}.
\bibitem[{Mou et~al.(2023)Mou, Wang, Xie, Wu, Zhang, Qi, Shan, and
  Qie}]{mou2023t2iadapter}
\bibinfo{author}{C.~Mou}, \bibinfo{author}{X.~Wang}, \bibinfo{author}{L.~Xie},
  \bibinfo{author}{Y.~Wu}, \bibinfo{author}{J.~Zhang}, \bibinfo{author}{Z.~Qi},
  \bibinfo{author}{Y.~Shan}, \bibinfo{author}{X.~Qie},
  \bibinfo{title}{T2i-adapter: Learning adapters to dig out more controllable
  ability for text-to-image diffusion models}, \bibinfo{year}{2023}.
  \href{http://arxiv.org/abs/2302.08453}{{\tt arXiv:2302.08453}}.
\bibitem[{von Platen et~al.(2022)von Platen, Patil, Lozhkov, Cuenca, Lambert,
  Rasul, Davaadorj, Nair, Paul, Berman, Xu, Liu, and
  Wolf}]{huggingface_diffusers}
\bibinfo{author}{P.~von Platen}, \bibinfo{author}{S.~Patil},
  \bibinfo{author}{A.~Lozhkov}, \bibinfo{author}{P.~Cuenca},
  \bibinfo{author}{N.~Lambert}, \bibinfo{author}{K.~Rasul},
  \bibinfo{author}{M.~Davaadorj}, \bibinfo{author}{D.~Nair},
  \bibinfo{author}{S.~Paul}, \bibinfo{author}{W.~Berman},
  \bibinfo{author}{Y.~Xu}, \bibinfo{author}{S.~Liu}, \bibinfo{author}{T.~Wolf},
  \bibinfo{title}{Diffusers: State-of-the-art diffusion models},
  \bibinfo{howpublished}{\url{https://github.com/huggingface/diffusers}},
  \bibinfo{year}{2022}.
\bibitem[{Xu et~al.(2023)Xu, Liu, Wu, Tong, Li, Ding, Tang, and
  Dong}]{xu2023imagereward}
\bibinfo{author}{J.~Xu}, \bibinfo{author}{X.~Liu}, \bibinfo{author}{Y.~Wu},
  \bibinfo{author}{Y.~Tong}, \bibinfo{author}{Q.~Li},
  \bibinfo{author}{M.~Ding}, \bibinfo{author}{J.~Tang},
  \bibinfo{author}{Y.~Dong}, \bibinfo{title}{Imagereward: Learning and
  evaluating human preferences for text-to-image generation},
  \bibinfo{year}{2023}. \href{http://arxiv.org/abs/2304.05977}{{\tt
  arXiv:2304.05977}}.
\bibitem[{N. et~al.(2021)N., I., CrowdSpeech, and DIY}]{toloka}
\bibinfo{author}{P.~N.}, \bibinfo{author}{S.~I.}, \bibinfo{author}{U.~D.
  CrowdSpeech}, \bibinfo{author}{V.~DIY}, \bibinfo{title}{Benchmark dataset for
  crowdsourced audio transcription}, \bibinfo{year}{2021}.
  \bibinfo{note}{Https://toloka.ai/}.
\bibitem[{Barratt and Sharma(2018)}]{barratt_note_2018}
\bibinfo{author}{S.~Barratt}, \bibinfo{author}{R.~Sharma}, \bibinfo{title}{A
  {Note} on the {Inception} {Score}}, \bibinfo{year}{2018}.
  \bibinfo{note}{ArXiv:1801.01973 [cs, stat]}.
\bibitem[{Obukhov et~al.(2020)Obukhov, Seitzer, Wu, Zhydenko, Kyl, and
  Lin}]{obukhov2020torchfidelity}
\bibinfo{author}{A.~Obukhov}, \bibinfo{author}{M.~Seitzer},
  \bibinfo{author}{P.-W. Wu}, \bibinfo{author}{S.~Zhydenko},
  \bibinfo{author}{J.~Kyl}, \bibinfo{author}{E.~Y.-J. Lin},
  \bibinfo{title}{High-fidelity performance metrics for generative models in
  pytorch}, \bibinfo{year}{2020}. \URLprefix
  \url{https://github.com/toshas/torch-fidelity}.
  \DOIprefix\doi{10.5281/zenodo.4957738}, \bibinfo{note}{version: 0.3.0, DOI:
  10.5281/zenodo.4957738}.
\bibitem[{Yang and Newsam(2010)}]{yi_ucmerced_2010}
\bibinfo{author}{Y.~Yang}, \bibinfo{author}{S.~Newsam},
  \bibinfo{title}{Bag-of-visual-words and spatial extensions for land-use
  classification}, \bibinfo{year}{2010}.
\bibitem[{Hessel et~al.(2021)Hessel, Holtzman, Forbes, Le~Bras, and
  Choi}]{hesselclipscore2021}
\bibinfo{author}{J.~Hessel}, \bibinfo{author}{A.~Holtzman},
  \bibinfo{author}{M.~Forbes}, \bibinfo{author}{R.~Le~Bras},
  \bibinfo{author}{Y.~Choi},
\newblock \bibinfo{title}{Clipscore: A reference-free evaluation metric for
  image captioning},
\newblock \bibinfo{journal}{Proceedings of the 2021 Conference on Empirical
  Methods in Natural Language Processing}  (\bibinfo{year}{2021}).
\bibitem[{Radford et~al.(2021)Radford, Kim, Hallacy, Ramesh, Goh, Agarwal,
  Sastry, Askell, Mishkin, Clark, Krueger, and Sutskever}]{radford_clip_2021}
\bibinfo{author}{A.~Radford}, \bibinfo{author}{J.~W. Kim},
  \bibinfo{author}{C.~Hallacy}, \bibinfo{author}{A.~Ramesh},
  \bibinfo{author}{G.~Goh}, \bibinfo{author}{S.~Agarwal},
  \bibinfo{author}{G.~Sastry}, \bibinfo{author}{A.~Askell},
  \bibinfo{author}{P.~Mishkin}, \bibinfo{author}{J.~Clark},
  \bibinfo{author}{G.~Krueger}, \bibinfo{author}{I.~Sutskever},
\newblock \bibinfo{title}{Learning transferable visual models from natural
  language supervision},
\newblock \bibinfo{journal}{CoRR} \bibinfo{volume}{abs/2103.00020}
  (\bibinfo{year}{2021}). \href{http://arxiv.org/abs/2103.00020}{{\tt
  arXiv:2103.00020}}.
\bibitem[{Zhengwentai(2023)}]{taited2023CLIPScore}
\bibinfo{author}{S.~Zhengwentai}, \bibinfo{title}{{clip-score: CLIP Score for
  PyTorch}},
  \bibinfo{howpublished}{\url{https://github.com/taited/clip-score}},
  \bibinfo{year}{2023}. \bibinfo{note}{Version 0.1.1}.
\bibitem[{Kolchinski et~al.(2019)Kolchinski, Zhou, Zhao, Gordon, and
  Ermon}]{Kolchinski_2019}
\bibinfo{author}{Y.~A. Kolchinski}, \bibinfo{author}{S.~Zhou},
  \bibinfo{author}{S.~Zhao}, \bibinfo{author}{M.~L. Gordon},
  \bibinfo{author}{S.~Ermon},
\newblock \bibinfo{title}{Approximating human judgment of generated image
  quality},
\newblock \bibinfo{journal}{CoRR} \bibinfo{volume}{abs/1912.12121}
  (\bibinfo{year}{2019}). \href{http://arxiv.org/abs/1912.12121}{{\tt
  arXiv:1912.12121}}.
\bibitem[{Srivastava(2023)}]{srivastavaImitationGameQuantifying2023}
\bibinfo{author}{A.~e.~a. Srivastava}, \bibinfo{title}{Beyond the {Imitation}
  {Game}: {Quantifying} and extrapolating the capabilities of language models},
  \bibinfo{year}{2023}. \bibinfo{note}{ArXiv:2206.04615 [cs, stat]}.
\bibitem[{Shi et~al.(2020)Shi, Wang, and Fang}]{shi_socialgood_2020}
\bibinfo{author}{Z.~R. Shi}, \bibinfo{author}{C.~Wang},
  \bibinfo{author}{F.~Fang},
\newblock \bibinfo{title}{Artificial intelligence for social good: {A} survey},
\newblock \bibinfo{journal}{CoRR} \bibinfo{volume}{abs/2001.01818}
  (\bibinfo{year}{2020}). \href{http://arxiv.org/abs/2001.01818}{{\tt
  arXiv:2001.01818}}.
\bibitem[{Floridi et~al.(2020)Floridi, Cowls, King, and Taddeo}]{Floridi2020}
\bibinfo{author}{L.~Floridi}, \bibinfo{author}{J.~Cowls},
  \bibinfo{author}{T.~C. King}, \bibinfo{author}{M.~Taddeo},
\newblock \bibinfo{title}{How to design ai for social good: Seven essential
  factors},
\newblock \bibinfo{journal}{Science and Engineering Ethics}
  \bibinfo{volume}{26} (\bibinfo{year}{2020}) \bibinfo{pages}{1771–1796}.
  \DOIprefix\doi{10.1007/s11948-020-00213-5}.
\bibitem[{Athalye et~al.(2018)Athalye, Engstrom, Ilyas, and
  Kevin}]{athalyeSynthesizingRobustAdversarial2018}
\bibinfo{author}{A.~Athalye}, \bibinfo{author}{L.~Engstrom},
  \bibinfo{author}{A.~Ilyas}, \bibinfo{author}{K.~Kevin},
\newblock \bibinfo{title}{Synthesizing robust adversarial examples},
\newblock in: \bibinfo{booktitle}{35th {Int}. {Conf}. {Mach}. {Learn}. {ICML}
  2018}, volume~\bibinfo{volume}{1}, \bibinfo{year}{2018}, pp.
  \bibinfo{pages}{449--468}. \URLprefix \url{http://arxiv.org/abs/1707.07397}.
\bibitem[{Z\"{u}ger and Asghari(2022)}]{zueger2022}
\bibinfo{author}{T.~Z\"{u}ger}, \bibinfo{author}{H.~Asghari},
\newblock \bibinfo{title}{Ai for the public. how public interest theory shifts
  the discourse on ai},
\newblock \bibinfo{journal}{AI \& SOCIETY} \bibinfo{volume}{38}
  (\bibinfo{year}{2022}) \bibinfo{pages}{815–828}.
  \DOIprefix\doi{10.1007/s00146-022-01480-5}.

\end{thebibliography}


\end{document}